\PassOptionsToPackage{table}{xcolor}
\documentclass[11pt]{article}
\usepackage{slt_arxiv}

\makeatletter
\newcommand{\earlymark}[1]{\begingroup
  \ifnum#1>0 \slt@slot{sltamber}\else\slt@off\fi\kern0.25ex
  \ifnum#1>1 \slt@slot{sltamber}\else\slt@off\fi\kern0.25ex
  \ifnum#1>2 \slt@slot{sltamber}\else\slt@off\fi\kern0.25ex
  \ifnum#1>3 \slt@slot{sltamber}\else\slt@off\fi
  \endgroup}
\makeatother
\newcommand{\linklogo}[1]{\raisebox{-0.28ex}{\includegraphics[height=1.95ex]{#1}}\kern0.32em}
\newenvironment{propbox}[1][Proposition]%
  {\begin{sltbox}[title={#1},unbreakable]{sltlink}\setlength{\parskip}{3pt}}{\end{sltbox}}
\newenvironment{takeawaybox}[1][Key Takeaway]%
  {\begin{sltbox}[title={#1},unbreakable]{sltamber}\setlength{\parskip}{3pt}}{\end{sltbox}}

\title{Late Entry in Looped Language Models}

\author{Ziqi Jin\authormark{1}, Jiaxi Li\authormark{2}, Weiwen Xu\authormark{3,\dagger},
        Aixin Sun\authormark{1,\dagger}}
\authornote{1}{Nanyang Technological University.}
\authornote{2}{Singapore University of Technology and Design.}
\authornote{3}{The Chinese University of Hong Kong.}
\correspondence{\textbf{Co-corresponding authors:} Weiwen Xu (\href{mailto:weiwen.xuu@gmail.com}{\texttt{weiwen.xuu@gmail.com}}) and Aixin Sun (\href{mailto:axsun@ntu.edu.sg}{\texttt{axsun@ntu.edu.sg}}).}
\runningtitle{Late Entry in Looped Language Models}
\runninghead{Preprint}

\begin{document}
\maketitle

\vspace{-12pt}
\begin{center}
\footnotesize
\linklogo{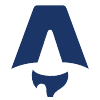}\textbf{Project:}~\href{https://xalp.net/late-entry/}{\texttt{xalp.net/late-entry}}
\;\;$\cdot$\;\;
\linklogo{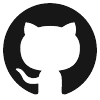}\textbf{Code:}~\href{https://github.com/Xalp/LateComer}{\texttt{github.com/Xalp/LateComer}}
\;\;$\cdot$\;\;
\linklogo{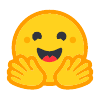}\textbf{Models:}~\href{https://huggingface.co/collections/Xalphinions/late-entry}{\texttt{hf.co/Xalphinions}}
\end{center}
\vspace{-4pt}

\begin{abstract}
Looped language models deepen reasoning at a compact parameter count by iterating a
shared layer stack $T$ times per token, turning recurrence depth directly into
sequential decoding latency. Conventional \emph{early-exit} decoding halts after
initial passes without cache completion, truncating cached key-value states of past
tokens to shallow depth. We study the opposite strategy, \textbf{late entry}, where new
tokens skip initial passes and enter directly into the final $K < T$ passes over a deep
context cache. Once lightweight fine-tuning resolves hidden-state norm mismatch at the
entry boundary, deep cached context substitutes for skipped token-local passes, allowing
late entry to outperform matched early exit across model families and pass budgets. To
turn this asymmetry into wall-clock speedups without stalling to build deep caches,
candidate blocks verify through only the final $K$ passes on the critical path while
previously committed tokens complete passes $1 \to T$ alongside subsequent steps at
\textbf{zero extra serial calls}. Projecting these finished $T$-pass wake states inside
the existing output GEMM (\textbf{Wake verification}, governed by logit-margin tolerance
$\tau \in [0, \infty]$) unifies \textbf{greedy-exact $T$-pass decoding} ($\tau = 0$,
matching $K = T$ block decoding within \texttt{bf16} reduction order), margin-gated
verification ($\tau > 0$), and non-rollback lossy late entry ($\tau = \infty$) inside a
single static CUDA graph. On Ouro-2.6B ($T = 4$), $K = 2$ Wake reaches \textbf{5.2$\times$}
greedy-exact speedup at $\tau = 0$ (\textbf{+26\%} over $K = 4$ block decoding on GSM8K)
and \textbf{6.8$\times$} at $\tau = \infty$ (\textbf{+16.9 points} over trained early exit
at matched speed). Scaling test-time recurrence across $T \in \{8, 16, 32, 64\}$ on
Huginn-3.5B cuts $K = T/4$ rollbacks \textbf{21.5$\times$} with depth, driving
greedy-exact ($\tau = 0$) speedups to \textbf{6.1$\times$} ($T = 32$) and
\textbf{8.4$\times$} ($T = 64$, \textbf{+153\%} over $K = 64$ block) and $\tau = \infty$
speedups to \textbf{10.6 to 12.5$\times$}.
\end{abstract}

\vspace{-2pt}
\begin{figure}[H]
\begin{center}
\includegraphics[width=0.95\textwidth]{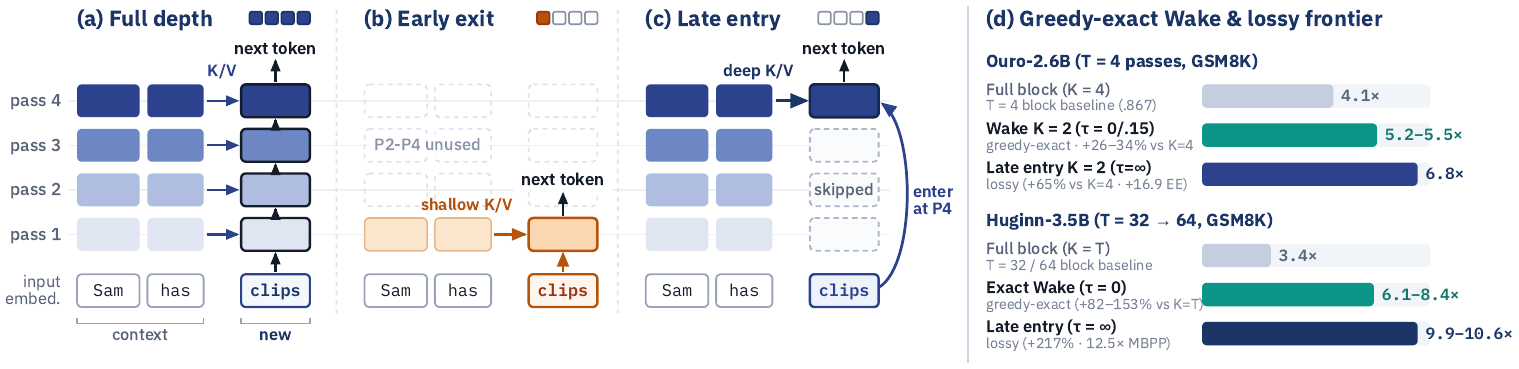}
\end{center}
\vspace{-8pt}
\caption{\textbf{Early exit vs.\ late entry in looped language models.}
\textbf{(a)~Full depth (\loopmark{4}):} runs all $T = 4$ passes.
\textbf{(b)~Early exit (\earlymark{1}):} halts after pass~$1$, truncating context KV
caches to shallow depth. \textbf{(c)~Late entry (\loopmark{1}):} skips passes~$1$
to~$3$ on the new token (\texttt{clips}) and enters pass~$4$ over a full-depth context
cache completed in parallel (Section~\ref{sec:method}). \textbf{(d)~Unified Wake and
late-entry frontier (GSM8K, $B = 8$, single H100):} reusing completed $T$-pass wake
states lifts throughput beyond the $K = T$ block baseline with greedy-exact fidelity at
$\tau = 0$ (\textbf{5.2$\times$}, \textbf{+26\%} on Ouro-2.6B; \textbf{6.1 to
8.4$\times$}, \textbf{+82\% to +153\%} on Huginn-3.5B across $T = 32 \to 64$) and
reaches \textbf{6.8$\times$} and \textbf{10.6$\times$} (\textbf{12.5$\times$} on MBPP)
at $\tau = \infty$.}
\label{fig:grids}
\vspace{-6pt}
\end{figure}

\section{Introduction}
\label{sec:intro}

Looped language models scale test-time compute without increasing model size by passing
each token repeatedly through a shared Transformer stack for $T$ recurrent passes
\citep{dehghani2019universal,geiping2025scaling,ouro2025}. During autoregressive
decoding, however, running all $T$ passes per generated token turns recurrence depth
directly into sequential latency. When each new token is restricted to $K < T$ serial
passes, \emph{which passes should it execute?} As shown in Figure~\ref{fig:grids},
conventional \textbf{early exit} without cache completion \citep{schuster2022calm} runs
the first $K$ passes and halts (Figure~\ref{fig:grids}b), leaving past key-value (KV)
caches truncated at shallow depth $K$ so all future tokens attend to immature context
states. We explore the opposite choice, \textbf{late entry}, which skips initial passes
and feeds each new token's embedding directly into the final $K$ passes over a deep
context KV cache (Figure~\ref{fig:grids}c).

At first glance, skipping initial passes seems destined to fail: on off-the-shelf
Ouro-1.4B and Ouro-2.6B ($T = 4$) \citep{ouro2025}, feeding raw embeddings into a late
pass breaks predictions immediately because late passes expect normalized hidden states
from earlier passes. Once lightweight fine-tuning aligns late passes to accept raw
embeddings, however, a clear asymmetry emerges between token depth and context depth:
one final pass over a four-pass context matches full-depth predictions more closely than
two initial passes over a two-pass context ($+10+$ accuracy points in sequential
decoding). Under identical training data, hyperparameters, and steps, spending $K$
serial passes at the end of the recurrence consistently outperforms spending them at the
beginning across both Ouro scales ($T = 4$; Section~\ref{sec:finding}) and $32$-pass
middle-recurrent Huginn-3.5B ($T = 32$; Section~\ref{sec:huginn}), showing that deep
cached context retains much of the benefit of recurrent depth.

Turning this finding into wall-clock speedups requires building deep context states
without stalling generation. In \textbf{late-entry block decoding} paired with
\textbf{Wake verification} (Section~\ref{sec:method}), parameter-free candidate blocks
\citep{liu2025tidar,jin2026mars} execute only the final $K$ passes on the critical path
to emit a \emph{tentatively accepted} frontier and provisional KV states, while earlier
tentative tokens complete true passes~$1$ through~$T$ as piggybacked batch rows at
\textbf{zero extra serial calls} (using spare per-call batch capacity in memory-bound
decoding rather than reducing FLOPs). Because that wake finishes exact causal $T$-pass
states $L = \lceil T / K \rceil$ steps later, projecting them inside the same output
GEMM (\textbf{Wake verification}, Section~\ref{sec:wake}) promotes matching tokens into
a \emph{permanently verified} $T$-pass prefix and rolls back rare mismatches, unifying
greedy-exact ($\tau = 0$), margin-gated ($0 < \tau < \infty$), and lossy ($\tau =
\infty$) decoding in one CUDA graph.

\begin{takeawaybox}[Key Takeaway]
In looped language models, deep cached context substitutes for token-local early passes.
Whereas existing looped speculative decoders \citep{ha2026wavefront,cho2026loopspec,li2026das,liu2026loopahead}
accelerate \emph{drafting} via shallow initial passes while keeping full $T$-pass
verification on every accepted token, \textbf{Late Entry + Wake Verification} compresses
\emph{critical-path verification itself} to the final $K < T$ passes over a pipelined
full-depth cache and reuses completed $1 \to T$ wake states in the same output GEMM to
preserve $T$-pass greedy equivalence ($\tau = 0$) or trade controlled margin tolerance
($\tau > 0$) for higher throughput.
\end{takeawaybox}

Our study is organized around three concrete contributions:
\begin{itemize}[leftmargin=1.5em,itemsep=2pt,topsep=2pt]
  \item \textbf{Pass-allocation asymmetry and error localization.} Across Ouro-1.4B,
    Ouro-2.6B ($T = 4$), and Huginn-3.5B ($T = 32$), adapted late entry consistently
    outperforms matched early exit; \texttt{fp32} interventions trace this gain to
    pass-coherent context $(K, V)$ pairs and show that $K < T$ discrepancies concentrate
    inside the $6\%$ uncertain tail (Section~\ref{sec:finding}).
  \item \textbf{Unified Late Entry + Wake Verification pipeline.} We pair $K$-pass
    critical-path block verification with cross-step $1 \to T$ cache completion and
    same-GEMM Wake verification, proving exact $T$-pass greedy equivalence and
    finite-step termination at $\tau = 0$ while spanning margin-gated ($\tau > 0$) and
    lossy ($\tau = \infty$) regimes in one engine (Proposition~1;
    Sections~\ref{sec:method} and~\ref{sec:wake}).
  \item \textbf{Scaling across benchmarks, recurrence depths, and hardware load.} Across
    $21{,}403$ examples on Ouro-2.6B ($T = 4$) and $T \in \{8, 16, 32, 64\}$ ($36 \to 260$
    effective layers) on Huginn-3.5B, Wake verification delivers \textbf{5.2$\times$} to
    \textbf{8.4$\times$} greedy-exact ($\tau = 0$) speedups and up to
    \textbf{12.5$\times$} lossy ($\tau = \infty$) speedups as $K = T/4$ rollbacks drop
    $21.5\times$ with depth, while an $L_p$ Roofline law calibrated strictly on $K = 4$
    predicts unseen $K < 4$ multi-request throughput across $C \in [1, 32]$ at
    \textbf{8.3\% MAPE} (Sections~\ref{sec:experiments} and~\ref{sec:roofline}).
\end{itemize}

\section{Early Exit vs.\ Late Entry: Where Should Passes Be Spent?}
\label{sec:finding}

Before building an inference system, we isolate a central pass-allocation question
under idealized context conditions: when each new token is limited to $K < T$ serial
passes, is it better to spend them on the initial passes or on the final passes? We
study this question on Ouro-1.4B ($24$ Transformer layers) and Ouro-2.6B ($48$
Transformer layers), two pretrained looped language models that iterate a shared
layer stack for $T = 4$ passes with pass-specific KV caches and untied input-output
embeddings \citep{ouro2025}. For clarity, Sections~\ref{sec:finding}
and~\ref{sec:method} develop our analysis and pipeline on $T = 4$ Ouro before
Section~\ref{sec:huginn} scales both to $32$-pass middle-recurrent Huginn-3.5B
($T = 32$).

In free autoregressive generation, a single altered prediction changes every
subsequent token, confounding single-step accuracy with trajectory drift. To measure
per-token prediction quality cleanly, we first evaluate \emph{teacher-forced
agreement} (Appendix~\ref{app:recipe}). We generate reference completions greedily at
full depth ($T = 4$), replay the exact same token prefixes under a reduced-pass
configuration, and record the fraction of positions whose greedy next-token
prediction matches full depth.

\subsection{Making Late Entry Feasible}
\label{sec:mismatch}

What happens if we apply late entry directly to an off-the-shelf looped language
model without any adaptation? We test the most extreme case on released Ouro-1.4B and
Ouro-2.6B: preceding context tokens retain their full four-pass KV caches, while the
current token skips passes~$1$ to~$3$ and feeds its raw token embedding directly into
pass~$4$ ($K = 1$). As shown in the first row of Table~\ref{tab:twosizes}, prediction
agreement with full depth drops to $.552$ on Ouro-1.4B and $.751$ on Ouro-2.6B, well
below simply truncating the entire model to pass~$1$ ($.938$ and $.949$).

\begin{table}[!t]
\caption{\textbf{Diagnosing raw late-entry failure on off-the-shelf purely looped
checkpoints.} Each row reports GSM8K teacher-forced next-token agreement with
full-depth ($T = 4$) decoding across $78{,}693$ token positions on Ouro-1.4B and
$70{,}101$ positions on Ouro-2.6B. In the top three rows, the current token executes
only pass~$4$ ($K = 1$) over a four-pass context cache; the bottom row runs pass~$1$
only for both token and context. A closed-form linear map recovers $60\%$ to $70\%$
of the gap to the pass-3 oracle without changing a single model weight.}
\label{tab:twosizes}
\begin{center}
\small
\begin{tabular}{lcc}
\toprule
Input state supplied to pass~4 ($K = 1$ over four-pass context) & Released Ouro-1.4B & Released Ouro-2.6B \\
\midrule
Raw token embedding \emph{(unadapted late entry)}            & .552 & .751 \\
Closed-form linear projection of the token embedding         & .816 & .919 \\
True pass-3 hidden state \emph{(diagnostic oracle ceiling)}  & .991 & .992 \\
\midrule
\emph{Reference:} 1-pass early exit (pass~1 only for all tokens) & .938 & .949 \\
\bottomrule
\end{tabular}
\end{center}
\end{table}

Inspecting internal activations reveals why raw late entry fails: pass~$1$ expects
raw token embeddings (median $\ell_2$ norm $0.76$), whereas pass~$4$ expects
layer-normalized hidden states emitted by pass~$3$ (norms $37$ to $51$). To test
whether this failure reflects an input representation mismatch rather than missing
early-pass computation, we fit a closed-form linear projection from raw token
embeddings to pass-4 inputs on held-out states while keeping all model weights frozen.
This linear probe alone raises agreement to $.816$ on Ouro-1.4B and $.919$ on
Ouro-2.6B (Table~\ref{tab:twosizes}), recovering \textbf{60\% and 70\%} of the gap to
the diagnostic oracle that supplies the true pass-3 state ($.991$ and $.992$).

Still, frozen input adapters on an unadapted model plateau at or below one-pass
truncation ($\le .949$; extended adapter diagnostics in Appendix~\ref{app:entry}).
To align late passes with raw embeddings directly, we fine-tune Ouro on
Dolci-Instruct-SFT \citep{olmo2025olmo3} in two one-epoch stages
(Appendix~\ref{app:recipe}): Stage~1 runs standard full-depth instruction SFT, and
Stage~2 exposes half of the training segments directly to raw token embeddings at pass
$T - K + 1$ ($K \sim \mathrm{Uniform}\{1, 2, 3\}$) over a full-depth prefix cache
(jointly formatted for parallel block decoding in Section~\ref{sec:blockverify}; full
recipe in Appendix~\ref{app:recipe}). Ablations in Appendix~\ref{app:arms} confirm that
late-entry pass exposure itself resolves the mismatch: Stage~2 lifts plain-embedding
late-entry agreement to \textbf{.962} on Ouro-2.6B, compared with $.699$ after Stage~1
alone and $.738$ under identical Stage-2 training without late-entry exposure. With the
input mismatch removed, we can now test whether deep context compensates for skipped
token passes.

\subsection{Deep Context Compensates for Skipped Token Passes}
\label{sec:decouple}

\begin{figure}[!t]
\begin{center}
\includegraphics[width=0.98\textwidth]{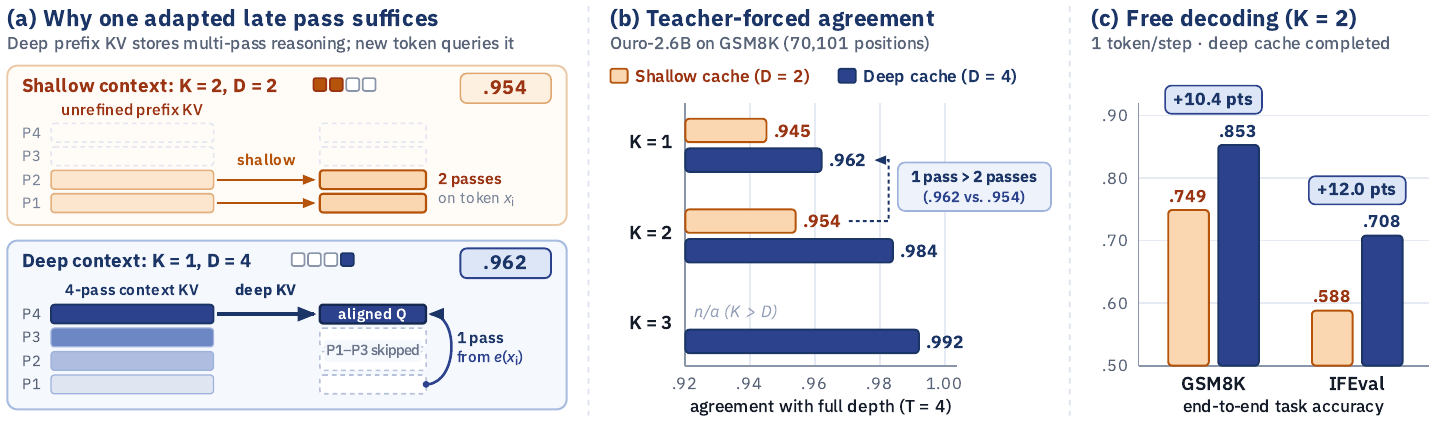}
\end{center}
\vspace{-4pt}
\caption{\textbf{Why a single adapted late pass suffices: deep context KV substitutes
for token-local passes on fine-tuned Ouro-2.6B.} \textbf{(a)~Mechanism:} under
shallow early exit ($K = 2, D = 2$, \earlymark{2}), token $x_i$ executes two passes
but attends only to unrefined two-pass prefix caches; under adapted late entry
($K = 1, D = 4$, \loopmark{1}), multi-pass contextual reasoning is already stored in
the four-pass prefix cache, allowing a single aligned pass~$4$ to query deep context
states directly from $e(x_i)$. \textbf{(b)~Teacher-forced decoupling:} varying
current-token passes $K$ (executing passes $D - K + 1, \dots, D$) and context depth
$D \in \{2, 4\}$ on GSM8K ($K = D = 4$ is $1.000$ by definition) reveals that one
pass over deep context (\textbf{.962}) beats two passes over shallow context
(\textbf{.954}). \textbf{(c)~Free decoding:} in single-token greedy generation at
$K = 2$, completing four-pass context caches lifts task accuracy by \textbf{+10.4
points} on GSM8K and \textbf{+12.0 points} on IFEval.}
\label{fig:ctxdepth}
\end{figure}

Why can a single late pass predict so accurately once the model is adapted? As
illustrated in Figure~\ref{fig:ctxdepth}a, passes~$1$ through~$T$ progressively
refine prefix representations and store multi-pass contextual reasoning inside the
deep KV cache $\mathcal{C}_{<i}^{(T)}$. Once Stage~2 aligns pass~$T$ to form valid
attention queries directly from a raw token embedding $e(x_i)$, the new token can
read out that distilled four-pass context in a single pass rather than re-deriving
depth locally, whereas early exit forces every attention head to read immature prefix
states (concurrently, \citet{li2026das} show in a training-free drafting setting that
pairing shallow initial-pass queries with full-depth prefix values improves draft
predictions). To test this mechanism quantitatively on fine-tuned Ouro-2.6B, we
independently vary two quantities under teacher forcing: the cached context depth
$D \in \{2, 4\}$, where all preceding tokens run passes $1, \dots, D$ to populate the
pass-aligned KV cache, and the current-token pass budget $K \le D$, where the current
token enters at pass $D - K + 1$ and executes passes $D - K + 1, \dots, D$. Under
this definition, $(K = 2, D = 2)$ runs passes~$1$ and~$2$ on both token and context
(standard two-pass early exit), whereas $(K = 1, D = 2)$ enters directly at pass~$2$
over a two-pass context.

Figure~\ref{fig:ctxdepth}b isolates each factor in turn. First, holding the current
token's budget fixed at $K = 2$ passes, deepening the context cache from $D = 2$ to
$D = 4$ raises agreement from $.954$ to $.984$. Second, holding a full-depth context
fixed at $D = 4$, reducing the current token from $K = 3$ ($.992$) to $K = 2$
($.984$) and even $K = 1$ ($.962$) retains high agreement. Together, these two
effects produce a clear cross-diagonal reversal: running just $K = 1$ pass over a
four-pass context (\textbf{.962}) outperforms running $K = 2$ passes over a two-pass
context (\textbf{.954}). Fine-tuned Ouro-1.4B exhibits the same reversal, reaching
$.948$ at $(K = 1, D = 4)$ versus $.922$ at $(K = 2, D = 2)$ with non-overlapping
bootstrap confidence intervals.

Small per-step differences compound substantially over free-running generation
(Figure~\ref{fig:ctxdepth}c). To isolate the end-to-end impact of context depth
without block drafting, we run diagnostic one-token-at-a-time greedy decoding at
$K = 2$ passes per token. Truncating every token and its cache to the first two
passes ($D = 2$) scores $.749$ on GSM8K and $.588$ on IFEval. Running the final two
passes per token while synchronously completing each emitted token's four-pass cache
before the next step ($D = 4$) lifts accuracy to \textbf{.853 (+10.4 points)} on
GSM8K and \textbf{.708 (+12.0 points)} on IFEval.

\begin{figure}[!t]
\begin{center}
\includegraphics[width=0.96\textwidth]{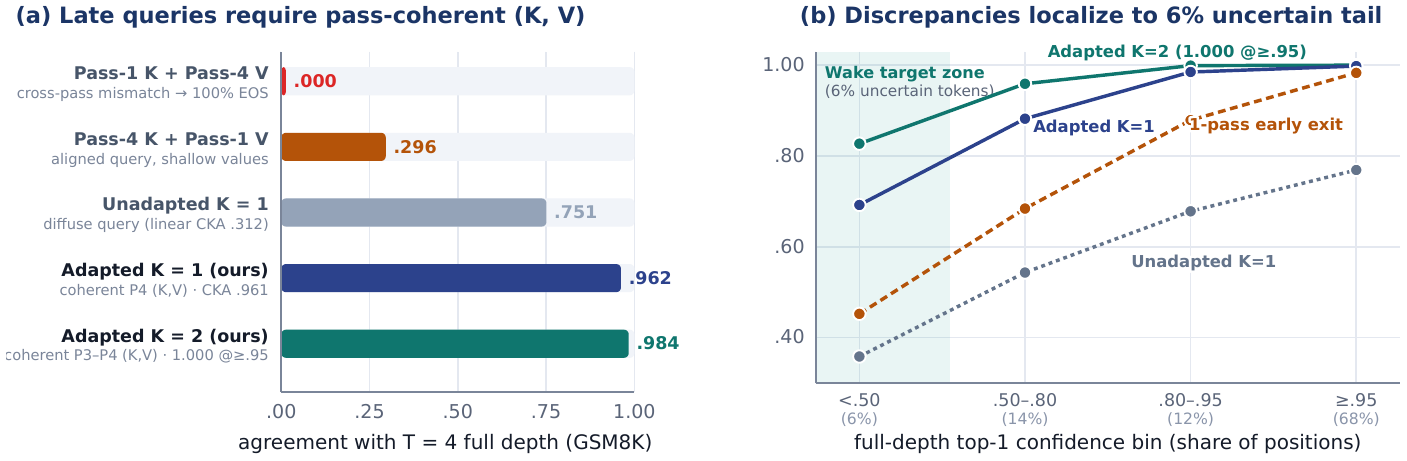}
\end{center}
\vspace{-6pt}
\caption{\textbf{Pass-coherent $(K, V)$ addressing and confidence localization on
Ouro-2.6B} (\texttt{fp32} probes across $30{,}736$ GSM8K positions).
\textbf{(a)~Cross-pass $(K, V)$ interventions:} swapping keys and values across passes
collapses agreement to $.000$ (Pass-1 $K$ + Pass-4 $V$) or $.296$ (Pass-4 $K$ + Pass-1
$V$), whereas Stage-2 alignment reaches linear CKA $.961$ with true pass-4 states and
$.962$ ($K = 1$) to $.984$ ($K = 2$) agreement over coherent deep $(K, V)$ caches.
\textbf{(b)~Confidence-binned agreement:} adapted late entry reaches \textbf{1.000}
($K = 2$) and $.998$ ($K = 1$) on the $68\%$ of confident tokens ($p_{\max} \ge .95$),
isolating residual discrepancies to the $6\%$ uncertain tail ($p_{\max} < .50$) targeted
by Wake verification.}
\label{fig:mechconf}
\vspace{-4pt}
\end{figure}

\paragraph{Why late passes work and where errors remain.}
Inspecting internal attention and hidden-state geometry in \texttt{fp32} across
$30{,}736$ GSM8K and $53{,}184$ IFEval positions (Figure~\ref{fig:mechconf}; full
attention-entropy, sink-mass, and distance metrics in Appendix~\ref{app:entry}) reveals
two structural properties. First, Stage-2 exposure aligns $K = 1$ hidden states directly
onto the pass-4 manifold (\textbf{linear CKA .961} vs.\ $.273$ to $.365$ unadapted), and
cross-pass KV interventions confirm that late queries require \emph{coherent same-pass
$(K, V)$ pairs} rather than deep values alone: pairing pass-1 keys with pass-4 values
collapses predictions to \texttt{<|endoftext|>} at $100\%$ of positions ($.000$
agreement), while pairing pass-4 keys with pass-1 values drops agreement to $.296$
(Figure~\ref{fig:mechconf}a; tokens more than $64$ positions back supply $53\%$ of
IFEval's total context-depth gain). Second, stratifying positions by full-depth top-1
confidence $p_{\max}$ (Figure~\ref{fig:mechconf}b) shows that on the \textbf{68\%} of
GSM8K tokens where full depth is confident ($p_{\max} \ge .95$), adapted late entry
already agrees at \textbf{.998} ($K = 1$) and \textbf{1.000} ($K = 2$), confining the
remaining $1.6\%$ discrepancy at $K = 2$ to the $6\%$ uncertain tail ($p_{\max} < .50$,
$.827$ agreement). This concentration motivates both halves of
Section~\ref{sec:method}: completing exact $1 \to T$ context caches in a cross-step
pipeline so generation never stalls (Section~\ref{sec:catchup}), and reusing those
finished wake states to catch rare uncertain-tail slips without slowing down the other
$98\%+$ of tokens (Section~\ref{sec:wake}).

\subsection{Late Entry vs.\ Early Exit Across Scales and Matched Fine-Tuning}
\label{sec:symprobe}

\begin{table}[!t]
\caption{\textbf{Same-checkpoint comparison of First $K$ versus Last $K$ passes
across scales and tasks.} Entries report teacher-forced next-token agreement with
full-depth ($T = 4$) decoding on identical Stage-2 checkpoints ($70{,}101$ GSM8K
positions on Ouro-2.6B; $28{,}939$ GSM8K and $61{,}723$ IFEval positions on
Ouro-1.4B). ``First $K$'' (\earlymark{2}) runs passes $1, \dots, K$ over a depth-$K$
context; ``Last $K$'' (\loopmark{2}) runs passes $T - K + 1, \dots, T$ over a
four-pass ($D = 4$) context.}
\label{tab:budget}
\begin{center}
\small
\setlength{\tabcolsep}{4.8pt}
\begin{tabular}{lcccccccc}
\toprule
& \multicolumn{4}{c}{\textbf{Ouro-2.6B} ($48$ layers, $T = 4$)} & \multicolumn{4}{c}{\textbf{Ouro-1.4B} ($24$ layers, $T = 4$)} \\
\cmidrule(lr){2-5}\cmidrule(lr){6-9}
& \multicolumn{2}{c}{GSM8K} & \multicolumn{2}{c}{IFEval} & \multicolumn{2}{c}{GSM8K} & \multicolumn{2}{c}{IFEval} \\
\cmidrule(lr){2-3}\cmidrule(lr){4-5}\cmidrule(lr){6-7}\cmidrule(lr){8-9}
Passes / token & First $K$ & Last $K$ & First $K$ & Last $K$ & First $K$ & Last $K$ & First $K$ & Last $K$ \\
\midrule
$K = 1$ \ (\earlymark{1} vs.\ \loopmark{1}) & .898 & \textbf{.962} & .757 & \textbf{.887} & .842 & \textbf{.948} & .743 & \textbf{.884} \\
$K = 2$ \ (\earlymark{2} vs.\ \loopmark{2}) & .954 & \textbf{.984} & .866 & \textbf{.943} & .922 & \textbf{.976} & .839 & \textbf{.941} \\
$K = 3$ \ (\earlymark{3} vs.\ \loopmark{3}) & .982 & \textbf{.992} & .949 & \textbf{.969} & .967 & \textbf{.990} & .938 & \textbf{.968} \\
\bottomrule
\end{tabular}
\end{center}
\end{table}

We first compare early-exit and late-entry pass placements on the \emph{exact same}
Stage-2 checkpoints across all reduced budgets $K \in \{1, 2, 3\}$, both model sizes,
and both GSM8K and IFEval. Table~\ref{tab:budget} compares truncating to the initial
$K$ passes (\emph{First $K$}, depth-$K$ context) against entering for the final $K$
passes (\emph{Last $K$}, four-pass context). Across all $12$ cells, spending passes
at the end beats spending them at the beginning on identical weights, with margins of
$+6.4$ to $+14.1$ points at $K = 1$. Note that each agreement score is measured
against that checkpoint's own full-depth output: although unadapted Ouro-2.6B showed
$.949$ one-pass self-agreement in Table~\ref{tab:twosizes} versus $.898$ here, the
unadapted base model achieves only $.729$ full-depth GSM8K accuracy ($.442$ on
IFEval; Appendix~\ref{app:arms}), lower than even two-pass truncated decoding on our
adapted model ($.747$ and $.597$).

Because Table~\ref{tab:budget} evaluates checkpoints exposed to late-entry blocks
during Stage~2, we next test whether giving early exit its own dedicated Stage-2
training closes the gap. Starting from the exact same Stage-1 checkpoint with identical
data order, optimization steps, and random seed, we train a matched early-exit
model whose reduced-budget blocks execute the \emph{first} $K$ passes instead of the
last $K$ (Appendix~\ref{app:recipe}). Dedicated early-exit training raises $K = 1$
truncation agreement on Ouro-2.6B from $.898$ to $.925$ on GSM8K and from $.757$ to
$.798$ on IFEval (while unexposed late entry stays at $.755$ and $.619$), yet it
still trails trained late entry by wide margins (\textbf{.962 vs.\
.925} on GSM8K and \textbf{.887 vs.\ .798} on IFEval), confirming across both identical
weights and matched fine-tuning that cached context depth outweighs token-local early
passes.

\section{Late-Entry Decoding}
\label{sec:method}

While Section~\ref{sec:finding} assumed idealized full-depth context states, a
practical decoder must construct those states for newly generated tokens without
stalling generation. Late-entry decoding achieves this by maintaining deep context
with a short pipeline lag across decoding steps. For a concrete mental model at
$T = 4$ and $K = 2$ (previewed in Figure~\ref{fig:step}), each decoding step performs
three coordinated actions:
\begin{enumerate}[leftmargin=1.6em,itemsep=2pt,topsep=2pt]
  \item \textbf{Tentative late-pass verification:} new candidate tokens skip passes~$1$
    and~$2$ and execute only passes~$3$ and~$4$ ($K = 2$ serial calls) to emit a
    \emph{tentatively accepted} token group.
  \item \textbf{Provisional KV exposure:} newly accepted tokens temporarily populate
    slots~$3$ and~$4$ with provisional KV states computed from their embeddings so
    the next step can proceed immediately without stalling.
  \item \textbf{Pipelined cache completion and Wake verification:} over the next two
    decoding steps, those tentative tokens ride along inside ongoing forward calls to
    compute true passes $1 \to 2$ and $3 \to 4$ at \textbf{zero extra serial calls},
    overwriting provisional KV entries and verifying finished $T$-pass states inside the
    same output GEMM to advance a \emph{permanently verified} $T$-pass prefix.
\end{enumerate}
We formalize looped pass mechanics in Section~\ref{sec:formulation}, describe
within-step block verification in Section~\ref{sec:blockverify}, detail cross-step
cache completion in Section~\ref{sec:catchup}, and unify greedy-exact and lossy decoding
via Wake verification in Section~\ref{sec:wake} (hardware Roofline and concurrency
scaling follow in Section~\ref{sec:roofline}).

\subsection{Looped Forward Pass and Late-Entry Formulation}
\label{sec:formulation}

Given a prefix $x_{\le i} = (x_1, \dots, x_i)$, let $e(x_i) \in \mathbb{R}^d$ denote
the input embedding of token $x_i$, and let $\mathcal{F}$ denote the shared stack of
Transformer layers iterated across passes $k \in \{1, \dots, T\}$. At pass $k$,
$\mathcal{F}$ maps the preceding normalized hidden state $\tilde{h}_i^{(k-1)}$
(initialized at $\tilde{h}_i^{(0)} = e(x_i)$) over the pass-aligned KV cache
$\mathcal{C}_{<i}^{(k)}$ of earlier positions $< i$:
\begin{equation}
  h_i^{(k)} = \mathcal{F}\!\left(\tilde{h}_i^{(k-1)};\, \mathcal{C}_{<i}^{(k)}\right),
  \qquad
  \tilde{h}_i^{(k)} = \mathrm{Norm}\!\left(h_i^{(k)}\right),
  \label{eq:recurrence}
\end{equation}
where $\mathrm{Norm}(\cdot)$ denotes layer normalization. When a token executes a
contiguous range of $K$ passes $\{a, \dots, b\} \subseteq \{1, \dots, T\}$, the model's
output readout maps the executed states $\tilde{h}_i^{(a)}, \dots, \tilde{h}_i^{(b)}$
through the unembedding projection $W_{\mathrm{out}}$ to next-token probabilities
$p_{a:b}(x_{i+1} \mid x_{\le i})$ without requiring hidden states from skipped passes
(via a pass-invariant stick-breaking gate normalized over $\{a, \dots, b\}$ on Ouro,
Appendix~\ref{app:recipe}, Equation~\eqref{eq:readout}; or directly through a
non-recurrent Coda on middle-recurrent Huginn, Section~\ref{sec:huginn}). Conventional
early exit runs $\{1, \dots, K\}$ (\earlymark{2}) and leaves $\mathcal{C}_{<i}^{(k)}$
empty for $k > K$. Late entry instead runs the final $K$ passes $\{T - K + 1, \dots, T\}$
(\loopmark{2}) of Equation~\eqref{eq:recurrence}, feeding $e(x_i)$ directly into pass
$T - K + 1$ while attending to context caches $\mathcal{C}_{<i}^{(k)}$ for
$k \in \{T - K + 1, \dots, T\}$.

Notice why pairing early exit with background cache completion does not yield a
distinct competitive baseline. Under Equation~\eqref{eq:recurrence}'s pass-aligned
routing, passes $1, \dots, K$ read \emph{only} $\mathcal{C}_{<i}^{(1)}, \dots,
\mathcal{C}_{<i}^{(K)}$ and never attend to $\mathcal{C}_{<i}^{(>K)}$, so completing
passes $K + 1 \to T$ in the background leaves Ouro's $K$-pass early-exit predictions
bit-for-bit unchanged (on middle-recurrent Huginn, where only the $2$-layer Coda reads
$\mathcal{C}_{<i}^{\mathrm{coda}}$, supplying an oracle $32$-pass cache lifts $8$-pass
early exit only from $.263$ to $.288$ vs.\ $.396$ for late entry;
Appendix~\ref{app:huginn}, Table~\ref{tab:hgsingle}). Conversely, rewiring those $K$
passes to query deep caches $\mathcal{C}_{<i}^{(T-K+1 \dots T)}$ instead of
$\mathcal{C}_{<i}^{(1 \dots K)}$ collapses algebraically into late entry itself because
all recurrent passes share identical weights $\mathcal{F}$ (including Ouro's
pass-invariant exit gate), and forfeiting true $\mathcal{C}_i^{(1 \dots K)}$ states on
the critical path means background completion must still compute all $1 \to T$ passes.

\subsection{Late-Pass Block Verification}
\label{sec:blockverify}

To emit multiple tokens per step and amortize cache completion cleanly, we use an
existing parameter-free mask-placeholder block decoder \citep{liu2025tidar,jin2026mars}
as our execution vehicle. We retain its single-call parallel verification and
lookahead layout unchanged, modifying only \emph{which recurrent passes execute} on
the critical path (passes $T - K + 1, \dots, T$ instead of $1, \dots, T$) and how
committed KV caches are completed afterward.

\begin{figure}[!t]
\begin{center}
\includegraphics[width=0.88\textwidth]{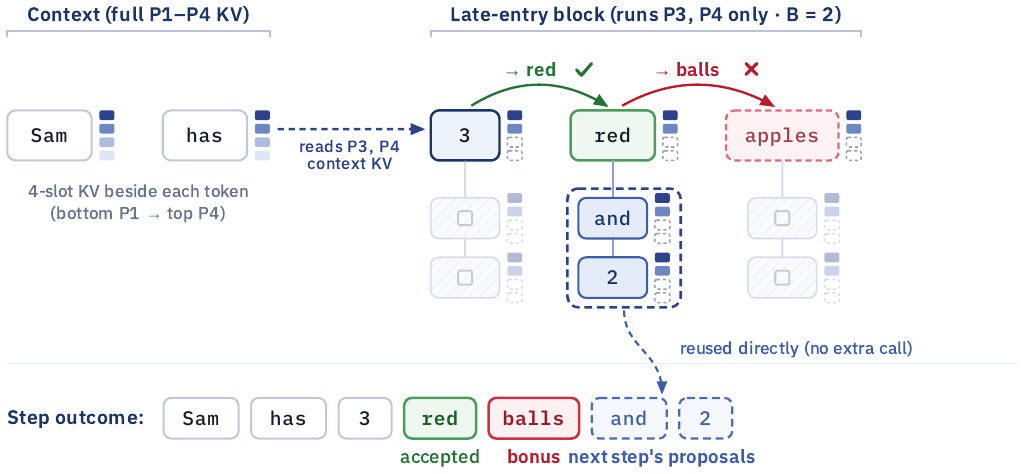}
\end{center}
\vspace{-4pt}
\caption{\textbf{Inside a single late-entry block step} ($T = 4, K = 2$, block size
$B = 2$). The last committed anchor token (``\texttt{3}'') and $B = 2$ candidate
proposals (``\texttt{red}'', ``\texttt{apples}'') each append a branch of $B = 2$
mask placeholders, forming $(1 + B)^2 = 9$ parallel rows that enter directly at
pass~$3$. Prefix verification accepts ``\texttt{red}'' ($\checkmark$), replaces the
rejected proposal with bonus token ``\texttt{balls}'' ($\times$), and reuses the
placeholder predictions behind ``\texttt{red}'' (``\texttt{and}'', ``\texttt{2}'') as
draft proposals to be verified from anchor ``\texttt{balls}'' in the next step.}
\label{fig:sched}
\end{figure}

Figure~\ref{fig:sched} illustrates one step for $T = 4, K = 2$, and block size
$B = 2$. Let $x_0$ denote the last committed anchor token (``\texttt{3}'') and let
$\hat{x}_1, \dots, \hat{x}_B$ denote $B$ candidate tokens proposed for the upcoming
positions (``\texttt{red}'', ``\texttt{apples}''). To draft the \emph{next} step's
proposals inside the same forward calls, we append a branch of $B$ mask-initialized
placeholders behind $x_0$ and behind each candidate $\hat{x}_m$, forming
$(1 + B)^2$ parallel rows in total. A structured causal mask restricts candidate
$\hat{x}_m$ to attend only to $x_0$ and $\hat{x}_{1:m}$ so verification never leaks
future tokens, while the $j$-th placeholder behind $\hat{x}_m$ attends to that prefix
plus its $j - 1$ preceding sibling placeholders to look $j$ positions ahead.

All $(1 + B)^2$ rows enter directly at pass $T - K + 1$ and execute only the final
$K$ passes ($K$ serial forward calls). After pass $T$, we evaluate the greedy
late-entry prediction $x^*_m = \arg\max p_{T-K+1:T}(\cdot \mid x_0, \hat{x}_{<m})$ at
each position. Let $c \in \{0, \dots, B\}$ be the length of the longest matching
prefix where $\hat{x}_{1:c} = x^*_{1:c}$. We \emph{tentatively accept} those $c$
proposals plus one bonus token $x^*_{c+1}$ (``\texttt{balls}'' in
Figure~\ref{fig:sched}), advancing the \emph{tentatively accepted frontier} by $1 + c$
tokens, and read the next step's $B$ draft proposals directly from the placeholder
branch behind accepted position $c$ with \textbf{zero separate drafting calls}. Because
that placeholder branch was evaluated alongside $\hat{x}_{c+1}$ rather than the newly
realized bonus token $x^*_{c+1}$, its outputs serve strictly as approximate draft
candidates; the next step sets $x^*_{c+1}$ as the new anchor $x_0$ and verifies those
proposals from scratch.

\subsection{Pipelined Cache Completion and Cost Profile}
\label{sec:catchup}

\begin{figure}[!t]
\begin{center}
\includegraphics[width=0.94\textwidth]{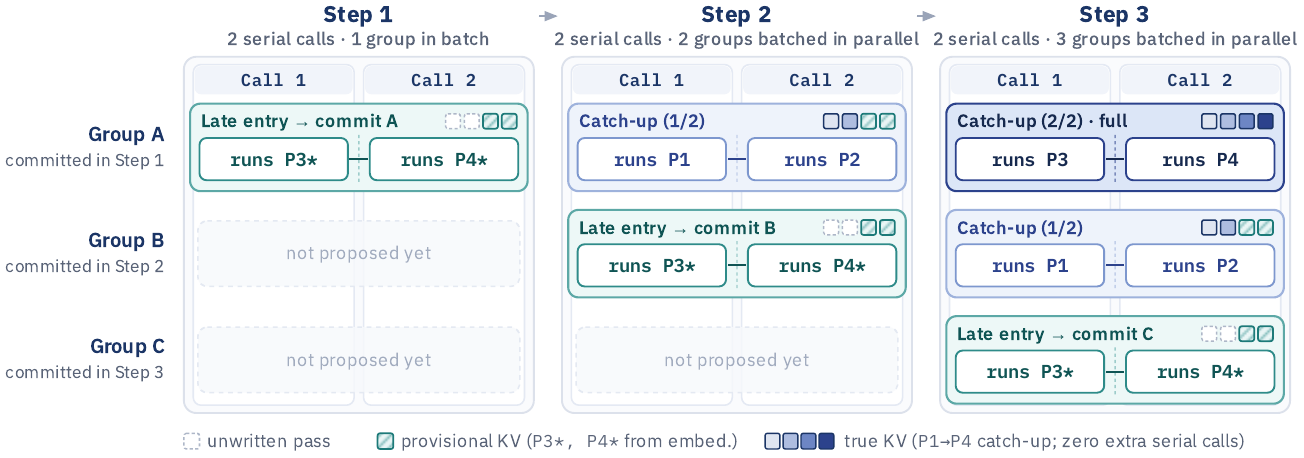}
\end{center}
\vspace{-4pt}
\caption{\textbf{Multi-step pipeline of late-entry decoding} ($T = 4, K = 2$). Each
step executes $K = 2$ serial forward calls. Tentatively accepted token group~A
($1 + c \le 1 + B$ tokens) runs late-entry passes~$3$ and~$4$ in Step~1 (writing
provisional KV states $\mathrm{P3}^*, \mathrm{P4}^*$), and then recomputes true passes
$1 \to 4$ as parallel batch rows inside the $K = 2$ calls of Step~2 (passes~$1, 2$)
and Step~3 (passes~$3, 4$), adding \textbf{zero extra serial forward calls}.}
\label{fig:step}
\end{figure}

When $1 + c$ tentatively accepted tokens advance the frontier after running only passes
$T - K + 1, \dots, T$, their KV cache slots for passes $1, \dots, T - K$ are still
empty, and slots $T - K + 1, \dots, T$ hold provisional keys and values computed from
raw embeddings ($\mathrm{P3}^*, \mathrm{P4}^*$ in Figure~\ref{fig:step}). Pausing
generation to recompute passes $1 \to T$ serially would add $T$ extra serial calls per
step and erase the latency savings of late entry.

Instead, we complete full-depth KV caches via cross-step pipelining
(Figure~\ref{fig:step}). Here, \emph{asynchronous} refers to delayed cache completion
across decoding steps inside the same GPU forward calls, rather than a separate
background CUDA stream. Because a looped language model shares identical layer
weights $\mathcal{F}$ across all passes, in-flight tentative token groups and the
active $(1 + B)^2$-row candidate block can execute different pass indices side by
side as parallel batch rows inside the same $K$ forward calls, each attending to its
own pass-aligned cache. Each tentative group advances through true passes $1 \to T$
at $K$ passes per step, finishing within $L = \lceil T / K \rceil$ steps ($2$ steps at
$T = 4, K = 2$, scaling to $4$ steps at $T = 32, K = 8$ in Section~\ref{sec:huginn};
when $K = T$, zero catch-up rows are scheduled).

Crucially, this schedule preserves exact causal KV dependencies. Because tentative
groups enter the wake queue in chronological order and advance $K$ passes per step,
whenever a group recomputes pass $k$, every earlier group has already written its
exact pass-$k$ KV states during a prior step. In Step~3 of Figure~\ref{fig:step}, for
example, Group~B computes passes~$1$ and~$2$ while attending to Group~A's true
pass-$1$ and pass-$2$ caches completed in Step~2. Meanwhile, active candidate blocks
briefly read provisional KV entries ($\mathrm{P3}^*, \mathrm{P4}^*$) for the one or
two most recent groups until true pass-$k$ states overwrite them.

Why does appending these catch-up rows add minimal wall-clock latency? It is
important to distinguish serial calls from arithmetic work: zero extra serial calls
does \emph{not} mean zero extra FLOPs, because tentatively accepted tokens execute $K$
verification passes plus $T$ wake passes. In single-sequence (batch-1) decoding,
however, GPU latency is memory-bandwidth bound rather than compute bound: once layer
weights $\mathcal{F}$ are streamed from memory for a forward call, extra token rows
incur little marginal latency (on Ouro-2.6B on an NVIDIA H100 GPU, forwarding $32$
parallel tokens takes only \textbf{9\% longer} than forwarding $1$ token). At $T = 4,
B = 8$, and $K = 2$, the active block has $(1 + 8)^2 = 81$ rows while at most two
groups of $\le 9$ tokens catch up simultaneously, adding at most $18$ rows
(\textbf{18\% of a 99-row batch}; all reported speeds include this work). Higher
request concurrency gradually saturates compute under a fixed large $B$ unless $B$ is
reduced alongside load (analyzed via our H100 Roofline law in
Section~\ref{sec:roofline} and Appendix~\ref{app:concurrency}). Finally,
cache-completion ablations in Appendix~\ref{app:catchup} verify both halves of this
design: reading provisional KV states for one or two steps costs \emph{zero measurable
accuracy} compared with synchronous before-next-step repair ($p = .28$ to $.77$),
whereas omitting cache completion altogether drops GSM8K accuracy at $K = 2$ by
\textbf{4.1 points} ($.839 \to .798$; Section~\ref{sec:huginn} and
Appendix~\ref{app:huginn} examine how multi-stage lag scales at $T = 32$).

\subsection{Wake Verification: Unifying Exact and Lossy Late Entry}
\label{sec:wake}

\begin{figure}[!t]
\begin{center}
\includegraphics[width=0.98\textwidth]{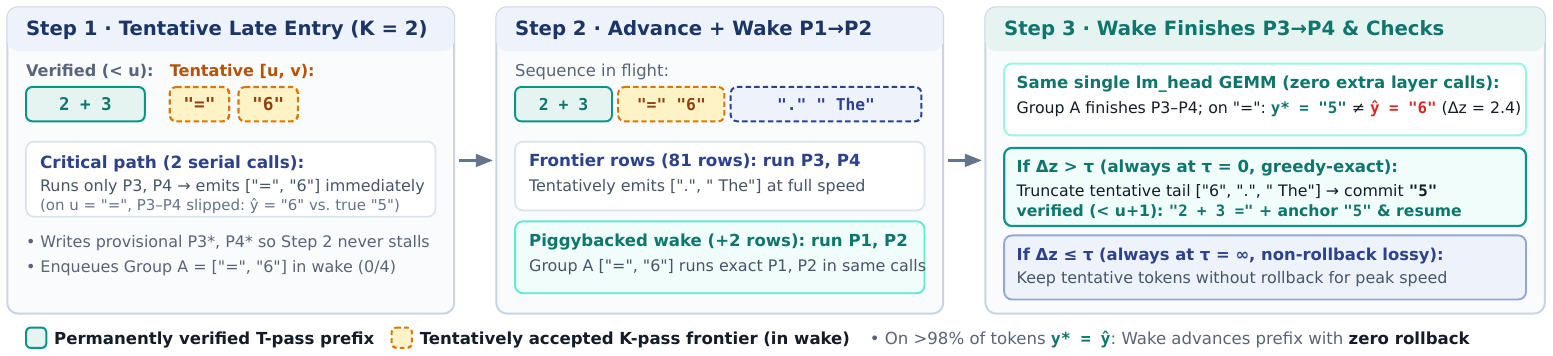}
\end{center}
\vspace{-6pt}
\caption{\textbf{Worked 3-step walkthrough of Wake verification and rollback} ($T = 4,
K = 2$). \textbf{Step~1:} over permanently verified prefix \texttt{"2 + 3"} ($< u$),
passes~$3$ and~$4$ tentatively accept Group~A $= [\texttt{"="}, \texttt{"6"}]$ at $[u,
v)$, slipping on position $u = \texttt{"="}$ ($\hat{y}_u = \texttt{"6"}$ instead of
\texttt{"5"}). \textbf{Step~2:} the frontier tentatively emits $[\texttt{"."},
\texttt{" The"}]$ while Group~A computes exact passes~$1$ and~$2$ inside the same
forward calls. \textbf{Step~3:} Group~A finishes passes~$3$ and~$4$ and projects inside
the \emph{same single} \texttt{lm\_head} \emph{GEMM}; detecting $y^*_u = \texttt{"5"} \neq
\texttt{"6"}$ ($\Delta z = 2.4$), $\tau = 0$ truncates the tentative tail beyond $u$,
retains exact KV states on \texttt{"2 + 3 ="} ($\le u$), commits \texttt{"5"} at $u + 1$,
and resumes.}
\label{fig:wake_walk}
\vspace{-4pt}
\end{figure}

Look closely at what pipelined cache completion in Section~\ref{sec:catchup} already
computes. At any step, generation partitions into a \textbf{permanently verified
$T$-pass prefix} ($< u$, holding exact $1 \to T$ KV states) followed by at most
$L = \lceil T / K \rceil$ in-flight groups along the \textbf{tentatively accepted
frontier}. Whenever the oldest tentative group $[u, v)$ ($v - u \le 1 + B$) finishes
pass~$T$ at the back of the wake, every earlier position $< u$ already holds its exact
$1 \to T$ KV cache and positions $i \in [u, v)$ have just executed exact causal passes
$1 \to T$. Hence the finished wake hidden states $\tilde{h}_i^{(1 \dots T)}$ (or Coda
states $\mathcal{F}_{\mathrm{coda}}(\mathrm{Norm}(s_i^{(T)}))$ on Huginn) are
\emph{mathematically identical} to sequential $T$-pass forward states. Whereas basic
lossy late entry discards those states after writing KV caches, \textbf{Wake
verification} projects them through $W_{\mathrm{out}}$ \emph{inside the exact same
single output GEMM} as the $(1 + B)^2$ active block rows ($90$ to $99$ rows vs.\ $81$
at $B = 8$, adding \textbf{zero Transformer layer calls, zero extra kernel launches,
and zero extra CUDA graphs}).

Figure~\ref{fig:wake_walk} walks through a concrete rollback at $T = 4, K = 2$. Suppose
prefix \texttt{"2 + 3"} ($< u$) is permanently verified. In \textbf{Step~1}, critical-path
passes~$3$ and~$4$ tentatively accept Group~A $= [\texttt{"="}, \texttt{"6"}]$ at
positions $[u, u + 2)$, where querying from $u = \texttt{"="}$ without passes~$1$
and~$2$ slipped and guessed $\hat{y}_u = \texttt{"6"}$. In \textbf{Step~2}, the frontier
advances at full speed with $[\texttt{"."}, \texttt{" The"}]$ while Group~A piggybacks
on the same calls to compute exact passes~$1$ and~$2$. In \textbf{Step~3}, Group~A
finishes passes~$3$ and~$4$ (writing exact $1 \to 4$ KV states on \texttt{"="}) and
projects inside the shared \texttt{lm\_head} GEMM, revealing exact $y^*_u = \texttt{"5"}
\neq \hat{y}_u = \texttt{"6"}$ with logit gap $\Delta z = 2.4$. On $> 98\%$ of positions
where $y^*_i = \hat{y}_i$, Group~A simply joins the permanently verified prefix with
zero rollback; when a mismatch exceeds tolerance $\tau$, Wake truncates the tentative
tail beyond $u$ and commits exact token \texttt{"5"} at $u + 1$.

Formally, let $z_{1:T}^{(i)} \in \mathbb{R}^{|\mathcal{V}|}$ denote the exact $T$-pass
logit vector at wake position $i \in [u, v)$, let $y^*_i = \arg\max_y z_{1:T}^{(i)}(y)$
be the true $T$-pass greedy next token, and let $\hat{y}_i = x_{i+1}$ be the token
tentatively accepted on the $K$-pass path. Parameterized by logit-margin tolerance
$\tau \ge 0$, Wake verification triggers a rollback at the earliest index
$i^* \in [u, v)$ satisfying
\begin{equation}
  y^*_{i^*} \neq \hat{y}_{i^*}
  \quad\text{and}\quad
  z_{1:T}^{(i^*)}\!\left(y^*_{i^*}\right) - z_{1:T}^{(i^*)}\!\left(\hat{y}_{i^*}\right) > \tau.
  \label{eq:wake}
\end{equation}
When Equation~\eqref{eq:wake} fires, the decoder truncates the tentative frontier beyond
$i^*$, commits exact $T$-pass token $y^*_{i^*}$ at position $i^* + 1$ (retaining all
exact KV caches $\le i^*$ already written by the wake), resets the wake queue to
$i^* + 1$, and resumes tentative $K$-pass decoding from anchor $y^*_{i^*}$. When the
tentative frontier emits \texttt{<eos>} or hits the token limit while wake groups remain
in flight, the decoder runs $\le L$ wake-drain steps before halting so premature stop
tokens are verified as well.

\begin{propbox}[Proposition 1: Exactness and Termination of $\tau = 0$ Wake Verification]
In exact arithmetic, Wake verification at $\tau = 0$ with EOS wake drain produces
\textbf{bit-for-bit identical token sequences} to sequential $T$-pass greedy decoding
for any $1 \le K \le T$ and $B \ge 0$, and terminates in at most $N_{\mathrm{max}}
+ L$ verified steps:
\begin{enumerate}[leftmargin=1.5em,itemsep=1.5pt,topsep=1.5pt]
  \item \emph{Inductive prefix exactness:} By induction on completed wake groups,
    every position $< u$ equals the sequential $T$-pass greedy prefix and holds exact
    $1 \to T$ KV states. Causal masking within $[u, v)$ ensures $y^*_u$ is exact; if
    $\hat{y}_i = y^*_i$ for all $i < m$, then $y^*_m$ is likewise exact. Hence either
    the entire group $[u, v)$ matches sequential $T$-pass decoding, or rollback fires
    at the exact first mismatch $i^* \in [u, v)$ and replaces $\hat{y}_{i^*}$ with the
    exact token $y^*_{i^*}$.
  \item \emph{Strict progress:} Every rollback advances the permanently verified
    $T$-pass prefix from $u$ to $i^* + 1 \ge u + 1 > u$, ruling out livelock, while
    wake drain verifies the final $\le L$ groups before termination (verified across
    $258/258$ adversarial \texttt{fp64} tests spanning divisible and non-divisible
    $T / K$ and premature-\texttt{<eos>} sequences).
\end{enumerate}
\end{propbox}

Varying $\tau$ along a single continuous knob unifies three regimes inside one engine:
\textbf{(i)~$\tau = 0$ (Exact Wake)} guarantees $T$-pass greedy equivalence in exact
arithmetic (and matches $K = T$ block decoding within \texttt{bf16} kernel reduction
order at $0.007\%$ to $0.09\%$ token discrepancy, below the $0.25\%$ to $0.33\%$
sequential \texttt{bf16} floor; Appendix~\ref{app:audit}) while running the critical
path at $K < T$ serial passes whenever tentative guesses agree ($> 98\%$ of positions;
Section~\ref{sec:decouple});
\textbf{(ii)~$0 < \tau < \infty$ (Margin-Gated Wake)} ignores benign near-tie
\texttt{bf16} or phrasing flips ($\Delta z \le \tau$) while rolling back confident
reasoning deviations ($\Delta z > \tau$); and
\textbf{(iii)~$\tau = \infty$} recovers non-rollback lossy late entry. Finally, because
Wake verification and late entry modify only \emph{which passes verify} a block, they
factorize cleanly from \emph{how candidate tokens are drafted}: pairing the identical
verifier with a second parameter-free $4/3/2$-gram context suffix drafter
\citep{fu2024lookahead,leviathan2023fast} alongside trained mask-placeholder proposals
directly tests whether the throughput multiplier from $K < T$ holds across distinct
acceptance regimes at zero extra batch rows or model parameters.

\section{Experiments}
\label{sec:experiments}

We organize our evaluation around four progressive questions:
\begin{enumerate}[leftmargin=1.6em,itemsep=1.5pt,topsep=2pt]
  \item \textbf{Pure pass-placement gain without rollback ($\tau = \infty$,
    Sections~\ref{sec:frontier} to~\ref{sec:huginn}):} Before activating Wake rollbacks,
    how much speedup and accuracy does non-rollback late entry achieve on its own across
    six benchmarks ($21{,}403$ items) on Ouro-2.6B ($T = 4$), against matched trained
    early exit and full-depth ($T = 4$) speculative decoders, and on $32$-pass
    middle-recurrent Huginn-3.5B ($T = 32$)?
  \item \textbf{Recovering greedy-exact fidelity ($\tau \in [0, \infty]$,
    Section~\ref{sec:exp_wake}):} When Wake verification reuses finished $1 \to T$ wake
    states inside the same output GEMM, how fast are greedy-exact ($\tau = 0$) and
    margin-gated ($\tau > 0$) decoding across models and candidate drafters?
  \item \textbf{Scaling with recurrence depth ($T \in \{8, 16, 32, 64\}$,
    Section~\ref{sec:tscale}):} Holding Huginn-3.5B weights fixed, how do Wake rollback
    rates and speedups scale as test-time loop depth grows from $36$ to $260$ effective
    layers?
  \item \textbf{Hardware Roofline and multi-request serving ($C \in [1, 32]$,
    Section~\ref{sec:roofline}):} Why do wake rows stay nearly free at batch~$1$, and
    how does throughput scale as concurrent load saturates GPU compute?
\end{enumerate}

Unless noted in Section~\ref{sec:roofline}, all runs use standard single-sequence
inference (batch size~$1$, greedy decoding, \texttt{bf16} precision, single NVIDIA H100
80GB GPU) across the full test sets of six benchmarks totaling $21{,}403$ examples:
\textbf{GSM8K} \citep[$1{,}319$ math word problems;][]{cobbe2021gsm8k},
\textbf{IFEval} \citep[$541$ strict instruction-following prompts;][]{zhou2023ifeval},
\textbf{MATH500} \citep[$500$ competition math problems;][]{hendrycks2021math,lightman2023letsverify},
\textbf{MBPP} \citep[$500$ Python synthesis tasks;][]{austin2021mbpp},
\textbf{BBH} \citep[$6{,}511$ multi-step reasoning problems;][]{suzgun2023bbh}, and
\textbf{MMLU-Pro} \citep[$12{,}032$ ten-choice questions;][]{wang2024mmlupro}, scored
via \texttt{lm-evaluation-harness} \citep{eval-harness} with gold-verified MATH500 and
MBPP extractors (Appendix~\ref{app:scorers}). Accuracy differences use two-sided paired
sign tests. In every single-engine table (Tables~\ref{tab:main}, \ref{tab:huginn},
\ref{tab:wake}, and~\ref{tab:tscale}), speedups and accuracy gaps are paired against
full-depth sequential decoding measured inside the same job on the same GPU (minor
sequential reference differences such as $.873$ at $63.4$~tokens/s in
Table~\ref{tab:main} vs.\ $.868$ at $62.3$~tokens/s in Table~\ref{tab:wake} reflect
$6/1{,}319$ \texttt{bf16} reduction-order ties across engine harnesses;
Appendix~\ref{app:audit}); in the cross-engine comparison of
Table~\ref{tab:headtohead}, all methods are normalized by the fastest sequential engine
across the three codebases ($62.4$ to $67.2$~tokens/s; Appendix~\ref{app:matched}) so
no external baseline is penalized by harness overhead.

\subsection{Accuracy and Throughput Across Six Benchmarks}
\label{sec:frontier}

\begin{figure}[!t]
\begin{center}
\includegraphics[width=0.94\textwidth]{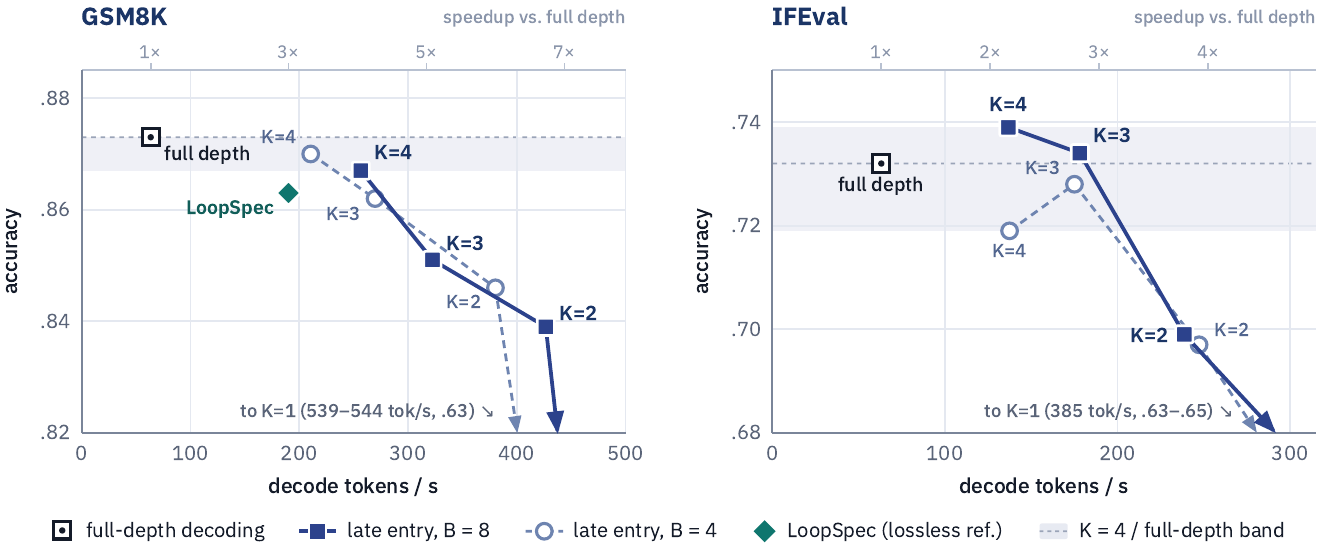}
\end{center}
\vspace{-6pt}
\caption{\textbf{Accuracy-throughput frontier of late-entry decoding on Ouro-2.6B}
across the full GSM8K ($n = 1{,}319$) and IFEval ($n = 541$) test sets for pass
budgets $K \in \{2, 3, 4\}$ and block sizes $B \in \{4, 8\}$ ($K = 1$ appears in
Figure~\ref{fig:contest} and Appendix~\ref{app:fullgrid}). LoopSpec marks its fastest
released lossless ($T = 4$) reference point on GSM8K.}
\label{fig:frontier}
\vspace{-4pt}
\end{figure}

Our total speedup combines two distinct mechanisms: lossless parallel block decoding
reduces the number of sequential decoding steps while still executing all $T = 4$
passes per step (captured by our $K = 4$ block baseline), whereas lossy late entry
further cuts the number of serial recurrent passes per step ($K = 3$ and $K = 2$) at a
controlled accuracy trade-off. Figure~\ref{fig:frontier} plots the
accuracy-throughput frontiers on GSM8K and IFEval across $K \in \{2, 3, 4\}$ and
block sizes $B \in \{4, 8\}$ (full grid in Appendix~\ref{app:fullgrid}). In
Figure~\ref{fig:frontier}, the $K = 4$ points serve as our lossless full-depth block
reference (minor accuracy differences from sequential decoding arise from \texttt{bf16}
reduction order; Appendix~\ref{app:audit}), and LoopSpec marks an external lossless
speculative reference point on GSM8K ahead of our six-benchmark boundary analysis in
Section~\ref{sec:headtohead}. Table~\ref{tab:main} reports both total speedups over
sequential decoding and incremental throughput gains over $K = 4$ block decoding at
$B = 8$.

\begin{table}[!t]
\caption{\textbf{Full-dataset accuracy and throughput on Ouro-2.6B ($T = 4$) across
six benchmarks at block size $B = 8$} ($21{,}403$ total items, single H100 GPU).
``Full depth'' is sequential $T = 4$ accuracy. ``Gap'' is accuracy minus full-depth
sequential accuracy in percentage points ($\dagger$ marks paired sign-test
$p < .05$). ``Speedup'' divides decode tokens/s by same-job full-depth sequential
tokens/s, while \textbf{(+X\%)} isolates the \emph{incremental throughput from late
entry} over $K = 4$ block decoding.}
\label{tab:main}
\begin{center}
\small
\setlength{\tabcolsep}{4.2pt}
\begin{tabular}{lrcrcrcrc}
\toprule
& & Full depth & \multicolumn{2}{c}{\loopmark{4}\;$K = 4$ (full)} & \multicolumn{2}{c}{\loopmark{3}\;$K = 3$ (late entry)} & \multicolumn{2}{c}{\loopmark{2}\;$K = 2$ (late entry)} \\
\cmidrule(lr){4-5}\cmidrule(lr){6-7}\cmidrule(lr){8-9}
Benchmark & $n$ & Acc. & Gap & Speedup & Gap & Speedup (vs.\ $K\!=\!4$) & Gap & Speedup (vs.\ $K\!=\!4$) \\
\midrule
GSM8K     & 1{,}319   & .873 & \cellcolor[HTML]{FBEBEA}$-$0.6 & 4.05$\times$ & \cellcolor[HTML]{F1B7B2}$-$2.2$^\dagger$ & 5.09$\times$\,\textbf{(+26\%)} & \cellcolor[HTML]{EA9088}$-$3.4$^\dagger$ & 6.73$\times$\,\textbf{(+66\%)} \\
IFEval    & 541       & .732 & \cellcolor[HTML]{E2F3EB}+0.7 & 2.16$\times$ & \cellcolor[HTML]{F7FCF9}+0.2 & 2.82$\times$\,\textbf{(+31\%)} & \cellcolor[HTML]{EA938C}$-$3.3$^\dagger$ & 3.78$\times$\,\textbf{(+75\%)} \\
MATH500   & 500       & .532 & \cellcolor[HTML]{FBEBEA}$-$0.6 & 4.06$\times$ & \cellcolor[HTML]{EEF8F3}+0.4 & 4.99$\times$\,\textbf{(+23\%)} & \cellcolor[HTML]{E67C73}$-$4.0$^\dagger$ & 6.49$\times$\,\textbf{(+60\%)} \\
MBPP      & 500       & .616 & \cellcolor[HTML]{D5EEE2}+1.0 & 3.65$\times$ & \cellcolor[HTML]{F7FCF9}+0.2 & 4.48$\times$\,\textbf{(+23\%)} & \cellcolor[HTML]{F6D1CE}$-$1.4 & 5.80$\times$\,\textbf{(+59\%)} \\
BBH       & 6{,}511   & .577 & \cellcolor[HTML]{FFFFFF}\phantom{+}0.0 & 3.30$\times$ & \cellcolor[HTML]{F7D4D2}$-$1.3$^\dagger$ & 4.30$\times$\,\textbf{(+30\%)} & \cellcolor[HTML]{EEA39D}$-$2.8$^\dagger$ & 6.16$\times$\,\textbf{(+87\%)} \\
MMLU-Pro  & 12{,}032  & .504 & \cellcolor[HTML]{FBFDFC}+0.1 & 3.49$\times$ & \cellcolor[HTML]{FDF5F4}$-$0.3 & 4.45$\times$\,\textbf{(+28\%)} & \cellcolor[HTML]{F5CBC7}$-$1.6$^\dagger$ & 5.91$\times$\,\textbf{(+69\%)} \\
\bottomrule
\end{tabular}
\end{center}
\vspace{-6pt}
\end{table}

At $K = 4$, every candidate block executes all four passes starting from pass~$1$
without skipping. Block decoding alone already accelerates generation by $2.16\times$
on IFEval and up to $4.05\times$ and $4.06\times$ on GSM8K and MATH500, with zero
statistically significant accuracy shift on any task ($0.01\%$ token discrepancy vs.\
a $0.27\%$ \texttt{bf16} floor on GSM8K; Appendix~\ref{app:audit}).

Reducing the critical path to $K = 3$ isolates the incremental contribution of late
entry on top of this block decoder. Skipping pass~$1$ boosts throughput by an
additional \textbf{23\% to 31\%} across all six benchmarks (reaching \textbf{2.82 to
5.09$\times$} total speedup, or $322.5$~tokens/s vs.\ $63.4$~tokens/s on GSM8K), while
accuracy changes by at most $0.4$ points on four of the six tasks (IFEval, MATH500,
MBPP, and MMLU-Pro) and trades $1.3$ points on BBH and $2.2$ points on GSM8K.

Skipping half the passes at $K = 2$ raises block throughput by \textbf{59\% to 87\%}
over $K = 4$ (\textbf{3.78 to 6.73$\times$} total speedup; $426.7$~tokens/s on GSM8K)
with bounded accuracy drops of $1.4$ to $4.0$ points. Dropping further to $K = 1$,
however, triggers a steep accuracy decline ($.629$ at $544.0$~tokens/s on GSM8K)
because a single pass both drafts and verifies its own proposals without cross-pass
refinement. Thus, $K = 3$ (near-full-depth quality) and $K = 2$ (high throughput)
define the practical operating regime of late-entry decoding.

\subsection{Controlled Comparison Against Early Exit}
\label{sec:contest}

Does the late-entry advantage from Section~\ref{sec:finding} carry over to end-to-end
block decoding through the identical $B = 8$ engine? We test this at two levels:
first on \emph{identical model weights} to isolate execution strategy alone, and
second across \emph{matched fine-tuned checkpoints} to evaluate each paradigm with
dedicated Stage-2 adaptation.

On identical weights, we take our adapted Ouro-2.6B checkpoint and compare running
its first $K$ passes (early-exit truncation) against running its last $K$ passes with
pipelined cache completion (late entry). At $K = 2$, truncating this checkpoint gives
the highest early-exit accuracy we measure on any model ($.747$ at $412$~tokens/s on
GSM8K and $.597$ on IFEval). Even on these exact same weights, late entry at $K = 2$
outperforms early-exit truncation by \textbf{+9.2 points} on GSM8K ($.839$ vs.\
$.747$) and \textbf{+10.2 points} on IFEval ($.699$ vs.\ $.597$), and leads by $+1.7$
and $+3.3$ points at $K = 3$.

\begin{table}[!t]
\caption{\textbf{End-to-end comparison of trained early exit and late
entry on Ouro-2.6B} under matched Stage-2 fine-tuning, decoded through the identical
$B = 8$ block engine on full test sets ($\dagger$ marks paired sign-test $p < .05$;
gaps computed prior to rounding). At
$K < 4$, early exit runs passes $1, \dots, K$ over a depth-$K$ cache, whereas late
entry runs passes $T - K + 1, \dots, T$ with pipelined full-depth cache completion.}
\label{tab:contest}
\begin{center}
\small
\setlength{\tabcolsep}{5pt}
\begin{tabular}{lcccccc}
\toprule
& \multicolumn{2}{c}{\textbf{Early exit, trained}} & \multicolumn{2}{c}{\textbf{Late entry, trained}} & & \\
\cmidrule(lr){2-3}\cmidrule(lr){4-5}
Passes / token & Acc. & tok/s & Acc. & tok/s & Gap (pts) & Sign-test $p$ \\
\midrule
\multicolumn{7}{l}{\emph{GSM8K} (1{,}319 test problems)} \\
$K = 1$ \ (\earlymark{1} vs.\ \loopmark{1}) & .330 & 657 & \textbf{.629} & 544 & \textbf{+29.9}$^\dagger$ & $4.4\times10^{-77}$ \\
$K = 2$ \ (\earlymark{2} vs.\ \loopmark{2}) & .669 & 431 & \textbf{.839} & 427 & \textbf{+16.9}$^\dagger$ & $2.7\times10^{-35}$ \\
$K = 3$ \ (\earlymark{3} vs.\ \loopmark{3}) & .785 & 314 & \textbf{.851} & 323 & \textbf{+6.5}$^\dagger$  & $8.3\times10^{-8}$ \\
$K = 4$ \ (full depth)                      & .822 & 246 & \textbf{.867} & 257 & \textbf{+4.5}$^\dagger$  & $2.2\times10^{-5}$ \\
\midrule
\multicolumn{7}{l}{\emph{IFEval} (541 test prompts)} \\
$K = 1$ \ (\earlymark{1} vs.\ \loopmark{1}) & .442 & 561 & \textbf{.645} & 386 & \textbf{+20.3}$^\dagger$ & $5.3\times10^{-16}$ \\
$K = 2$ \ (\earlymark{2} vs.\ \loopmark{2}) & .586 & 286 & \textbf{.699} & 239 & \textbf{+11.3}$^\dagger$ & $2.9\times10^{-7}$ \\
$K = 3$ \ (\earlymark{3} vs.\ \loopmark{3}) & .691 & 183 & \textbf{.734} & 178 & \textbf{+4.3}$^\dagger$  & $.023$ \\
$K = 4$ \ (full depth)                      & .721 & 138 & \textbf{.739} & 137 & \textbf{+1.8}\phantom{$^\dagger$} & $.34$ \\
\bottomrule
\end{tabular}
\end{center}
\vspace{-5pt}
\end{table}

\begin{figure}[!t]
\begin{center}
\includegraphics[width=0.88\textwidth]{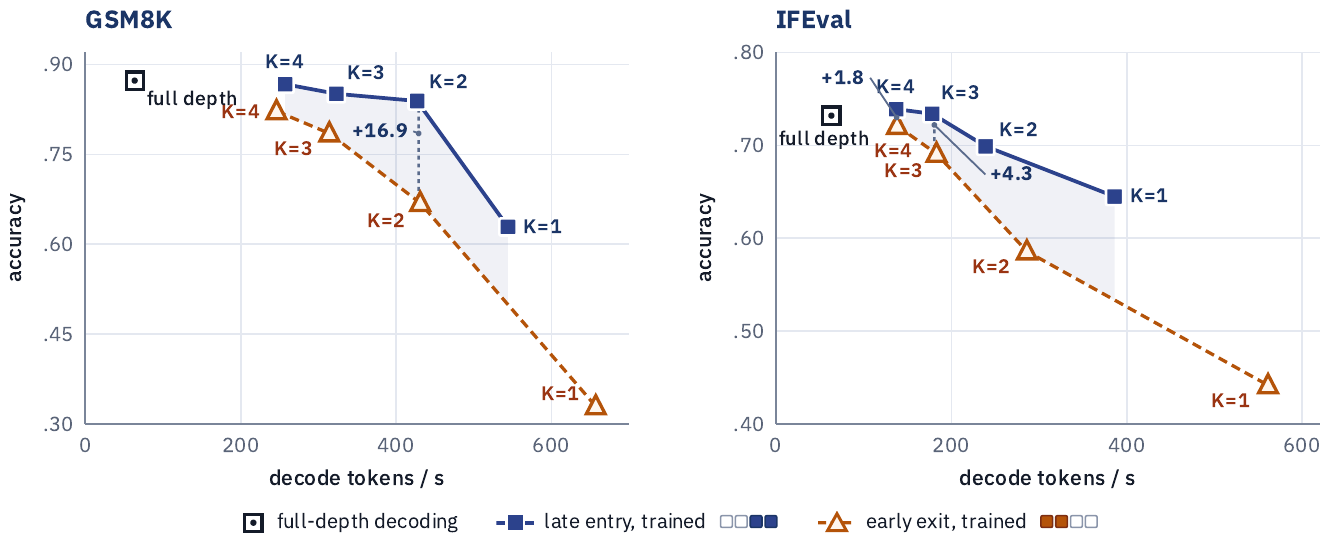}
\end{center}
\vspace{-6pt}
\caption{\textbf{Accuracy versus decode throughput for trained late entry and trained
early exit on Ouro-2.6B} at $B = 8$ (plotting Table~\ref{tab:contest}). Vertical
calipers mark matched-throughput operating points (\textbf{+16.9 points} at
${\sim}430$~tokens/s on GSM8K; \textbf{+4.3} and \textbf{+1.8 points} on IFEval).}
\label{fig:contest}
\vspace{-4pt}
\end{figure}

Next, Table~\ref{tab:contest} and Figure~\ref{fig:contest} compare our late-entry
checkpoint against the matched trained early-exit checkpoint across $K \in \{1,
2, 3, 4\}$. Notice first at $K = 4$ that forcing shallow passes to emit final
predictions during Stage~2 lowers the early-exit model's full-depth ceiling on GSM8K
($.822$ vs.\ $.867$). Even relative to each checkpoint's own $K = 4$ ceiling,
dropping from $K = 4$ to $K = 2$ costs late entry only $2.8$ points on GSM8K ($4.0$
on IFEval) compared with $15.2$ points for trained early exit ($13.5$ on IFEval;
additional adaptive-exit controls in Appendix~\ref{app:seqrepro}). Across all eight
cells, trained late entry dominates the overlapping speed range, leading by
\textbf{+16.9 accuracy points} at matched speed around $430$~tokens/s on GSM8K
($.839$ vs.\ $.669$ at $K = 2$) and by \textbf{+4.3 points} around $180$~tokens/s on
IFEval ($.734$ vs.\ $.691$ at $K = 3$).

\subsection{Comparison with Full-Depth ($T = 4$) Speculative Decoders}
\label{sec:headtohead}

To situate non-rollback late entry ($K < T, \tau = \infty$) relative to methods that
verify every token through all $T = 4$ passes, we evaluate two speculative decoders
tailored to looped language models across the \emph{full} test sets of all six
benchmarks ($21{,}403$ examples in total): \textbf{LoopSpec} \citep[official
implementation, fastest released configuration;][]{cho2026loopspec} and
\textbf{LoopAhead} \citep[best of three reproduced training arms;][]{liu2026loopahead}.
Because LoopSpec, LoopAhead, and our $K = 4$ block baseline verify every drafted token
through all $T = 4$ passes, they map out the throughput reachable under full-depth
critical-path verification. On each task, every method's decode tokens/s is normalized
by the \emph{fastest} measured full-depth sequential decoder across all three engines
($62.4$ to $67.2$~tokens/s; protocol details in Appendix~\ref{app:matched}).

\begin{table}[!t]
\caption{\textbf{Full-depth ($T = 4$) speculative decoders versus non-rollback
($\tau = \infty$) late entry on Ouro-2.6B} across all six benchmarks ($21{,}403$ total
items, single H100 GPU). Speedups divide each method's decode tokens/s by the fastest
measured full-depth sequential decoder on that benchmark. LoopSpec, LoopAhead, and our
$K = 4$ block baseline run full-depth $T = 4$ verification; $K = 3$ and $K = 2$
($\tau = \infty$) trade bounded accuracy shifts (small numbers report full-set accuracy,
with full-depth sequential accuracy beside each denominator) for higher throughput,
while Section~\ref{sec:exp_wake} activates $\tau = 0$ Wake to surpass $K = 4$ with
greedy-exact fidelity.}
\label{tab:headtohead}
\begin{center}
\small
\setlength{\tabcolsep}{3.2pt}
\begin{tabular}{lcccccc}
\toprule
& & \multicolumn{3}{c}{\textbf{Lossless $T = 4$ verification}} & \multicolumn{2}{c}{\textbf{Lossy late entry (ours)}} \\
\cmidrule(lr){3-5}\cmidrule(lr){6-7}
Benchmark & Full depth (tok/s) & LoopSpec & LoopAhead
  & \loopmark{4}\;Ours $K = 4$ & \loopmark{3}\;$K = 3$ & \loopmark{2}\;$K = 2$ \\
\midrule
GSM8K    & 67.2 {\scriptsize(.873)} & \cellcolor[HTML]{EBEFF7}2.84$\times$ & \cellcolor[HTML]{E5EBF5}2.94$\times$ & \cellcolor[HTML]{D4DDEE}3.82$\times$\,{\scriptsize.867} & \cellcolor[HTML]{A5B5DB}4.80$\times$\,{\scriptsize.851} & \cellcolor[HTML]{7A90C9}6.35$\times$\,{\scriptsize.839} \\
IFEval   & 64.7 {\scriptsize(.732)} & \cellcolor[HTML]{EBEFF7}2.34$\times$ & \cellcolor[HTML]{DCE4F2}2.67$\times$ & \cellcolor[HTML]{F1F4FA}2.12$\times$\,{\scriptsize.739} & \cellcolor[HTML]{B9C7E4}2.75$\times$\,{\scriptsize.734} & \cellcolor[HTML]{8DA2D2}3.68$\times$\,{\scriptsize.699} \\
MATH500  & 62.7 {\scriptsize(.532)} & \cellcolor[HTML]{E3E9F4}2.85$\times$ & \cellcolor[HTML]{ECEFF8}2.59$\times$ & \cellcolor[HTML]{CED8EC}4.11$\times$\,{\scriptsize.526} & \cellcolor[HTML]{9EAFD8}5.07$\times$\,{\scriptsize.536} & \cellcolor[HTML]{718AC6}6.63$\times$\,{\scriptsize.492} \\
MBPP     & 65.1 {\scriptsize(.616)} & \cellcolor[HTML]{E7ECF6}2.86$\times$ & \cellcolor[HTML]{E5EBF5}2.90$\times$ & \cellcolor[HTML]{D5DEEF}3.64$\times$\,{\scriptsize.626} & \cellcolor[HTML]{AEBCDE}4.51$\times$\,{\scriptsize.618} & \cellcolor[HTML]{859ACE}5.95$\times$\,{\scriptsize.602} \\
BBH      & 65.2 {\scriptsize(.577)} & \cellcolor[HTML]{ECEFF8}2.62$\times$ & \cellcolor[HTML]{D9E2F1}3.25$\times$ & \cellcolor[HTML]{D7E0F0}3.29$\times$\,{\scriptsize.577} & \cellcolor[HTML]{B2C0E0}4.35$\times$\,{\scriptsize.564} & \cellcolor[HTML]{7A91C9}6.33$\times$\,{\scriptsize.549} \\
MMLU-Pro & 62.4 {\scriptsize(.504)} & \cellcolor[HTML]{E4EAF5}2.75$\times$ & \cellcolor[HTML]{EEF1F8}2.44$\times$ & \cellcolor[HTML]{D4DDEE}3.56$\times$\,{\scriptsize.504} & \cellcolor[HTML]{ACBBDE}4.56$\times$\,{\scriptsize.501} & \cellcolor[HTML]{8096CC}6.12$\times$\,{\scriptsize.488} \\
\bottomrule
\end{tabular}
\end{center}
\end{table}

Table~\ref{tab:headtohead} separates full-depth verification from critical-path pass
reduction. Within \textbf{full-depth $T = 4$ verification}, LoopSpec ($2.34$ to
$2.86\times$), LoopAhead ($2.44$ to $3.25\times$), and our $K = 4$ block baseline
($2.12$ to $4.11\times$) map out the speed range reachable when every step executes all
four passes (with $K = 4$ leading on five tasks and trailing on IFEval). Reducing
critical-path verification to $K < 4$ under non-rollback late entry ($\tau = \infty$)
reaches \textbf{2.75 to 5.07$\times$} at $K = 3$ ($\le 0.4$-point accuracy change on
four tasks; $1.3$ and $2.2$ points on BBH and GSM8K) and \textbf{3.68 to 6.63$\times$}
at $K = 2$ ($1.4$ to $4.0$-point trade-offs). Moreover, full-$T$ critical-path
verification is \emph{not} the ceiling of greedy-exact decoding: as
Section~\ref{sec:exp_wake} demonstrates next, activating Exact Wake ($\tau = 0$) at
$K = 2$ also surpasses $K = 4$ block throughput (reaching $5.20\times$ on GSM8K and
$5.23\times$ on MATH500) while preserving greedy-exact $T$-pass predictions.

\subsection{Scaling to Deep Middle-Recurrence: Huginn-3.5B ($T = 32$)}
\label{sec:huginn}

To test whether late entry generalizes beyond shallow four-pass recurrence and
Ouro's stick-breaking readout, we evaluate \textbf{Huginn-3.5B}
\citep[\texttt{huginn-0125};][]{geiping2025scaling}, a middle-recurrent architecture
comprising a $2$-layer non-recurrent \emph{Prelude}, a $4$-layer \emph{Core} block
iterated $T = 32$ times via a learned input-injection adapter, and a $2$-layer
\emph{Coda} ($132$ effective Transformer layers per token, with only $8$ unique
parameter layers). Under late entry at budget $K < 32$, each candidate block executes
the $2$-layer Prelude once, skips the initial $32 - K$ Core passes to enter directly
at pass $32 - K + 1$ from Huginn's standard initialized latent state ($s_i^{(32-K)} = s_i^{(0)}$,
drawn from a fixed-seed truncated Gaussian) over
a $32$-pass context cache, and exits through the $2$-layer Coda; previously committed
tokens complete true Core passes $1 \to 32$ and Coda across $\lceil 32 / K \rceil$
subsequent steps without recomputing Prelude KV states (Appendix~\ref{app:huginn},
Equation~\eqref{eq:huginn}).
Following our Ouro protocol, we run one epoch of full-depth Stage-1 SFT on
Dolci-Instruct-SFT followed by matched $7{,}406$-step Stage-2 block fine-tuning
($K \in \{2, 4, 6, 8, 12, 16, 24\}$) for both late entry and a dedicated early-exit
control, and benchmark all sequential, late-entry, synchronous, and early-exit decoders
inside a unified CUDA-graphed paged FlashAttention-3 engine so wall-clock comparisons
are free of dispatch overhead (Appendix~\ref{app:huginn}). We evaluate the four
benchmarks in our suite on which the $8$-layer \texttt{huginn-0125} base model exhibits
non-trivial capability (GSM8K, IFEval, BBH, and MBPP; $8{,}871$ test items), omitting
ten-choice MMLU-Pro and competition-level MATH500 where the base model lacks task
capability and scores near chance. Notably, because Prelude
already maps raw embeddings into normalized contextual anchors $u_i$ and re-injects
$u_i$ at every Core step, Huginn provides a natural architectural validation of
Section~\ref{sec:mismatch}'s norm-mismatch diagnosis: even on the
\emph{early-exit-trained} checkpoint (which never saw late entry in Stage~2), $K = 8$
late entry beats that checkpoint's own trained $K = 8$ early exit on GSM8K ($.327$ vs.\
$.271$, $p = 2.2\times10^{-6}$), before dedicated late-entry training aligns Core to
$s_i^{(32-K)} = s_i^{(0)}$ and lifts accuracy to \textbf{.354}.

\begin{figure}[!t]
\begin{center}
\includegraphics[width=0.94\textwidth]{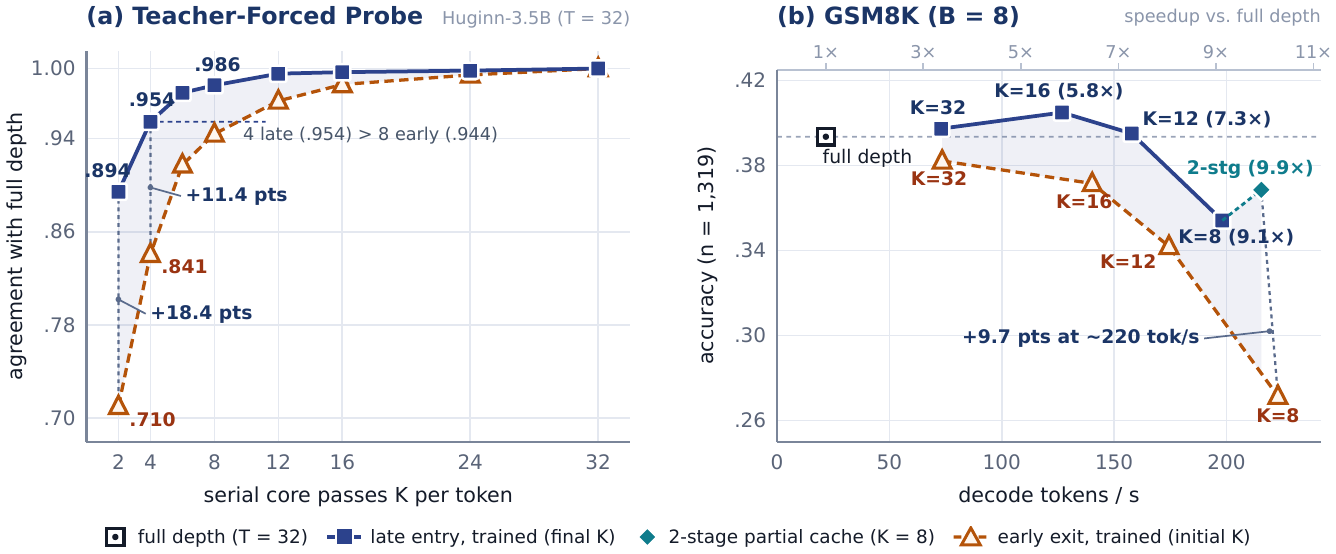}
\end{center}
\vspace{-6pt}
\caption{\textbf{Generalization to 32-pass middle-recurrent Huginn-3.5B} ($T = 32$,
matched $7{,}406$-step Stage-2 checkpoints). \textbf{(a)~Teacher-forced probe:}
across all pass budgets $K < 32$ on GSM8K, late entry over a $32$-pass context
outperforms trained early exit, with $4$ late passes (\textbf{.954})
exceeding $8$ trained early-exit passes (\textbf{.944}). \textbf{(b)~End-to-end GSM8K
accuracy-throughput frontier ($B = 8, n = 1{,}319$):} inside our unified CUDA-graphed
paged FlashAttention-3 engine, late entry outperforms trained early exit over a broad
range of practical operating points, reaching \textbf{5.84$\times$} ($126.8$~tokens/s,
\textbf{.405}) at $K = 16$, \textbf{7.27$\times$} ($157.8$~tokens/s, \textbf{.395}) at
$K = 12$, \textbf{9.13$\times$} ($198.1$~tokens/s, \textbf{.354}) at four-stage
$K = 8$, and \textbf{9.93$\times$} ($215.5$~tokens/s, \textbf{.369}) under a two-stage
partial-cache $K = 8$ variant (\textbf{+9.7 points} over trained early exit at
$\sim 220$~tokens/s).}
\label{fig:huginn}
\vspace{-4pt}
\end{figure}

\begin{table}[!t]
\caption{\textbf{Full-dataset accuracy and throughput on Huginn-3.5B ($T = 32$) across
all four capability-active benchmarks at block size $B = 8$} ($8{,}871$ total items,
single H100 GPU, unified CUDA-graphed paged FlashAttention-3 engine; definitions match
Table~\ref{tab:main}).}
\label{tab:huginn}
\begin{center}
\small
\setlength{\tabcolsep}{3.8pt}
\shrinktofit{%
\begin{tabular}{lrccrcrcrc}
\toprule
& & \multicolumn{2}{c}{Full depth ($T\!=\!32$)} & \multicolumn{2}{c}{$K = 32$ (full block)} & \multicolumn{2}{c}{$K = 16$ (late entry)} & \multicolumn{2}{c}{$K = 8$ (late entry)} \\
\cmidrule(lr){3-4}\cmidrule(lr){5-6}\cmidrule(lr){7-8}\cmidrule(lr){9-10}
Benchmark & $n$ & Acc. & tok/s & Gap & Speedup & Gap & Speedup (vs.\ $K\!=\!32$) & Gap & Speedup (vs.\ $K\!=\!32$) \\
\midrule
GSM8K    & 1{,}319  & .393 & 21.7 & \cellcolor[HTML]{EEF8F3}+0.4 & 3.37$\times$ & \cellcolor[HTML]{D1EDE0}+1.1 & 5.84$\times$\,\textbf{(+74\%)} & \cellcolor[HTML]{E77F76}$-$3.9$^\dagger$ & \phantom{0}9.13$\times$\,\textbf{(+171\%)} \\
IFEval   & 541      & .551 & 21.6 & \cellcolor[HTML]{FFFFFF}\phantom{+}0.0 & 2.93$\times$ & \cellcolor[HTML]{C0E6D3}+1.5 & 5.24$\times$\,\textbf{(+79\%)} & \cellcolor[HTML]{F9DCD9}$-$0.9\phantom{$^\dagger$} & \phantom{0}8.96$\times$\,\textbf{(+206\%)} \\
BBH      & 6{,}511  & .283 & 22.6 & \cellcolor[HTML]{F7FCF9}+0.2 & 3.79$\times$ & \cellcolor[HTML]{EEF8F3}+0.3 & 6.65$\times$\,\textbf{(+75\%)} & \cellcolor[HTML]{F9D7D5}$-$1.0$^\dagger$ & 10.86$\times$\,\textbf{(+186\%)} \\
MBPP     & 500      & .234 & 22.4 & \cellcolor[HTML]{F7FCF9}+0.2 & 4.46$\times$ & \cellcolor[HTML]{FDF5F4}$-$0.2 & 7.76$\times$\,\textbf{(+74\%)} & \cellcolor[HTML]{FBEBEA}$-$0.6\phantom{$^\dagger$} & 12.41$\times$\,\textbf{(+178\%)} \\
\bottomrule
\end{tabular}}
\end{center}
\vspace{-5pt}
\end{table}

Figure~\ref{fig:huginn} and Table~\ref{tab:huginn} confirm that all three properties
observed at $T = 4$ strengthen at $T = 32$. First, under teacher forcing
(Figure~\ref{fig:huginn}a), just $K = 4$ final Core passes ($1/8$ of the recurrence)
reach \textbf{.954} agreement with full $32$-pass predictions, surpassing $8$ trained
early-exit passes (\textbf{.944}), while single-token $K = 8$ late entry matches
$32$-pass GSM8K accuracy with zero degradation ($.396$ vs.\ $.396$, $+13.3$ points over
$8$-pass truncation; Appendix~\ref{app:huginn}, Table~\ref{tab:hgsingle}). Second,
whereas at $T = 4$ parallel block verification supplies the larger share of total
speedup, scaling to $T = 32$ ($21.6$ to $22.6$~tokens/s sequential) \emph{inverts this
balance} so late entry drives the majority of wall-clock gain (Table~\ref{tab:huginn}):
halving Core depth to $K = 16$ adds \textbf{+74\% to +79\%} over $K = 32$ block decoding
(\textbf{5.24 to 7.76$\times$} total) with zero statistically significant accuracy drop
on any task ($K = 12$ reaches $6.44$ to $9.75\times$ within $0.6$ points;
Appendix~\ref{app:huginn}), and cutting to $K = 8$ adds \textbf{+171\% to +206\%}
(\textbf{8.96 to 12.41$\times$} total, up to $278.3$~tokens/s on MBPP) with no
significant drop on IFEval or MBPP.

Across matched wall-clock speeds, late entry consistently dominates trained early exit
(Figure~\ref{fig:huginn}b; full grid in Table~\ref{tab:hgcontest}): on GSM8K, $K = 12$
late entry ($.395$ at $157.8$~tokens/s) is both faster and more accurate than $16$-pass
early exit ($.372$ at $140.3$~tokens/s), while $K = 8$ leads $8$-pass early exit by
\textbf{+8.3 points} ($p = 3.1\times10^{-10}$), and on IFEval late entry leads by
\textbf{+4.8 points} around $225$~tokens/s ($p = .029$). The sole non-rollback trade-off
at $K = 8$ on GSM8K ($-3.9$ points) stems from four-stage pipeline lag leaving three
recent blocks provisional: shrinking the lag via a two-stage partial-cache variant
($17 \to 32$) recovers half the gap at \textbf{9.93$\times$} (\textbf{+9.7 points} over
$8$-pass early exit; Table~\ref{tab:hgcatchup}), while activating Exact Wake ($\tau = 0$)
in Section~\ref{sec:exp_wake} eliminates this lag \emph{completely}.

\subsection{Unifying Lossless and Lossy Decoding via Wake Verification}
\label{sec:exp_wake}

\begin{figure}[!t]
\begin{center}
\includegraphics[width=0.95\textwidth]{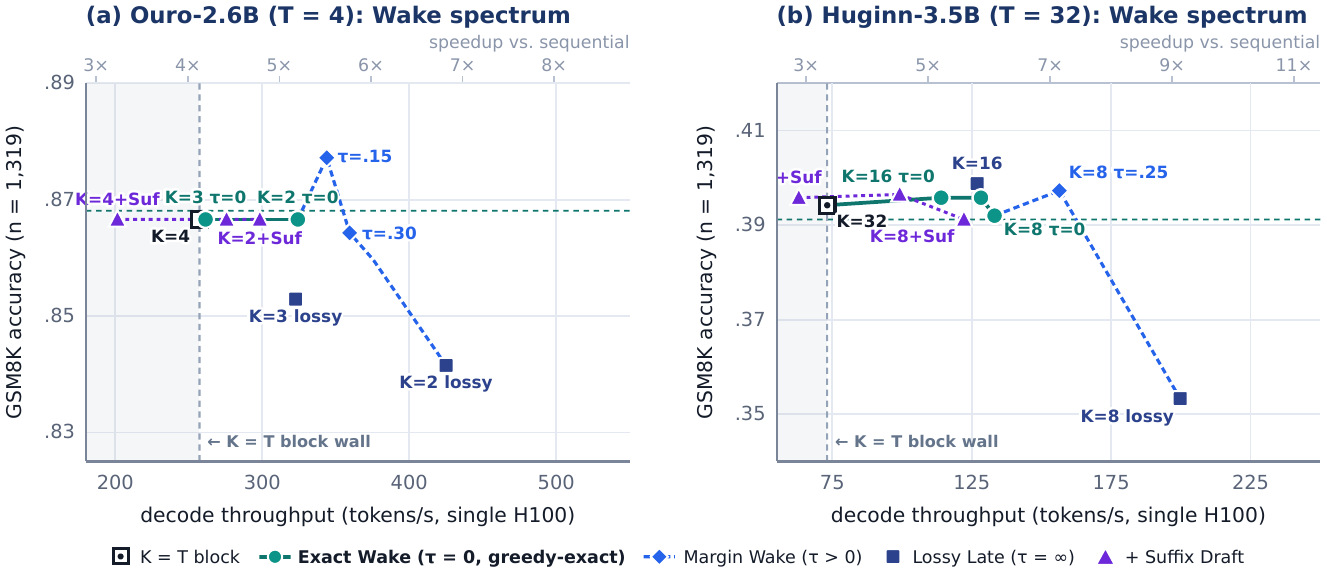}
\end{center}
\vspace{-6pt}
\caption{\textbf{Unified Wake verification spectrum ($\tau \in [0, \infty]$) and
drafter factorization on GSM8K ($n = 1{,}319, B = 8$).} \textbf{(a)~Ouro-2.6B
($T = 4$):} $\tau = 0$ Exact Wake at $K = 2$ lifts throughput \textbf{+26\%} beyond the
$K = 4$ block baseline ($4.13\times \to \mathbf{5.20\times}$) with identical $.867$
accuracy and $0.007\%$ committed-token discrepancy, while margin gating ($\tau \in
[0.15, 0.30]$) reaches \textbf{5.52 to 5.77$\times$} (\textbf{+34\% to +40\%}) at
$.877$ to $.864$ accuracy before connecting smoothly to $\tau = \infty$ ($6.82\times$).
\textbf{(b)~Huginn-3.5B ($T = 32$):} $\tau = 0$ Exact Wake eliminates four-stage
$K = 8$ lag completely, reaching \textbf{5.21$\times$} (\textbf{+56\%}) at $K = 16$,
\textbf{6.09$\times$} (\textbf{+82\%}) at $K = 8$, and \textbf{7.15$\times$}
(\textbf{+114\%}) at $\tau = 0.25$ with zero accuracy drop.}
\label{fig:wake}
\vspace{-4pt}
\end{figure}

Sections~\ref{sec:frontier} to~\ref{sec:huginn} established non-rollback late entry
($\tau = \infty$) as a high-speed lossy regime, leaving small accuracy trade-offs at
aggressive compression ($-2.7$ to $-4.0$ points at $K = 2$ on Ouro math tasks, and
$-3.9$ points under four-stage lag at $K = 8$ on Huginn GSM8K). We now evaluate
\textbf{Wake verification} (Section~\ref{sec:wake}), which projects the finished
$1 \to T$ cache-completion wake states inside the same output GEMM to sweep $\tau \in
[0, \infty]$ at zero extra layer calls. Figure~\ref{fig:wake} and Table~\ref{tab:wake}
report full-dataset accuracy, decode throughput, rollback rates per $100$ emitted
tokens (\texttt{rb}), and teacher-forced fidelity audits across Ouro-2.6B ($T = 4$) and
Huginn-3.5B ($T = 32$).

\begin{table}[!t]
\caption{\textbf{Wake verification spectrum ($\tau \in [0, \infty]$) across Ouro-2.6B
($T = 4$) and Huginn-3.5B ($T = 32$) at block size $B = 8$} (full test sets, single
H100 GPU). ``rb'' denotes Wake rollbacks per $100$ emitted tokens; \textbf{(+X\%)}
measures incremental throughput over same-job $K = T$ full-depth block decoding
($\dagger$ marks paired sign-test $p < .05$ vs.\ sequential $T$-pass decoding).}
\label{tab:wake}
\begin{center}
\small
\setlength{\tabcolsep}{3.5pt}
\shrinktofit{%
\begin{tabular}{llccrccrccrc}
\toprule
& & \multicolumn{1}{c}{Seq.\ $T$} & \multicolumn{2}{c}{$K = T$ Block} & \multicolumn{3}{c}{\textbf{Exact Wake ($\tau = 0$, greedy-exact)}} & \multicolumn{3}{c}{\textbf{Margin Wake ($\tau > 0$)}} & \multicolumn{1}{c}{Lossy ($\tau = \infty$)} \\
\cmidrule(lr){3-3}\cmidrule(lr){4-5}\cmidrule(lr){6-8}\cmidrule(lr){9-11}\cmidrule(lr){12-12}
Benchmark & Budget & Acc.\ {\scriptsize(tok/s)} & Acc. & Spd. & Acc. & rb & Spd.\ {\scriptsize(vs.\ $K\!=\!T$)} & Acc.\ {\scriptsize($\tau$)} & rb & Spd.\ {\scriptsize(vs.\ $K\!=\!T$)} & Acc.\ / Spd. \\
\midrule
\multicolumn{12}{l}{\emph{Ouro-2.6B ($T = 4$ passes; teacher-forced $\tau = 0$ audit diff: $0.007\%$ GSM8K, $0.09\%$ MATH500, $0.025\%$ MBPP)}} \\
GSM8K    & $K = 3$ & .868 {\scriptsize(62.3)} & .867 & 4.13$\times$ & \textbf{.867} & 1.14 & 4.19$\times$\,\textbf{(+2\%)}  & \textbf{.870} {\scriptsize(.25)} & 0.29 & \textbf{4.55$\times$\,(+10\%)} & .853 / 5.18$\times$ \\
GSM8K    & $K = 2$ & .868 {\scriptsize(62.3)} & .867 & 4.13$\times$ & \textbf{.867} & 2.02 & \textbf{5.20$\times$\,(+26\%)} & \textbf{.877} {\scriptsize(.15)} & 1.26 & \textbf{5.52$\times$\,(+34\%)} & .842$^\dagger$ / 6.82$\times$ \\
MATH500  & $K = 2$ & .532 {\scriptsize(62.7)} & .526 & 4.08$\times$ & \textbf{.526} & 2.11 & \textbf{5.23$\times$\,(+28\%)} & \textbf{.548} {\scriptsize(.30)} & 0.96 & \textbf{5.83$\times$\,(+43\%)} & .492$^\dagger$ / 6.61$\times$ \\
IFEval   & $K = 2$ & .732 {\scriptsize(63.8)} & .739 & 2.15$\times$ & \textbf{.739} & 5.98 & \textbf{2.66$\times$\,(+24\%)} & .719 {\scriptsize(.15)}          & 4.08 & \textbf{2.94$\times$\,(+36\%)} & .699$^\dagger$ / 3.76$\times$ \\
MBPP     & $K = 2$ & .616 {\scriptsize(63.4)} & .626 & 3.79$\times$ & \textbf{.626} & 2.00 & \textbf{4.61$\times$\,(+22\%)} & \textbf{.618} {\scriptsize(.30)} & 0.70 & \textbf{5.13$\times$\,(+35\%)} & .602 / 6.20$\times$ \\
\midrule
\multicolumn{12}{l}{\emph{Huginn-3.5B ($T = 32$ Core passes; unified CUDA-graphed paged FlashAttention-3 engine)}} \\
GSM8K    & $K = 16$ & .391 {\scriptsize(21.9)} & .394 & 3.35$\times$ & \textbf{.396} & 0.48 & \textbf{5.21$\times$\,(+56\%)} & \textbf{.404} {\scriptsize(.25)} & 0.05 & \textbf{5.42$\times$\,(+62\%)} & .399 / 5.80$\times$ \\
GSM8K    & $K = 8$  & .391 {\scriptsize(21.9)} & .394 & 3.35$\times$ & \textbf{.392} & 2.22 & \textbf{6.09$\times$\,(+82\%)} & \textbf{.397} {\scriptsize(.25)} & 0.88 & \textbf{7.15$\times$\,(+114\%)} & .353$^\dagger$ / 9.13$\times$ \\
IFEval   & $K = 16$ & .560 {\scriptsize(21.5)} & .547 & 2.91$\times$ & \textbf{.547} & 0.37 & \textbf{4.85$\times$\,(+66\%)} & \textbf{.560} {\scriptsize(.25)} & 0.02 & \textbf{5.10$\times$\,(+75\%)} & .562 / 5.21$\times$ \\
IFEval   & $K = 8$  & .560 {\scriptsize(21.5)} & .547 & 2.91$\times$ & \textbf{.556} & 2.03 & \textbf{6.46$\times$\,(+122\%)} & .543 {\scriptsize(.25)}          & 0.59 & \textbf{7.59$\times$\,(+161\%)} & .542 / 8.85$\times$ \\
MBPP     & $K = 16$ & .234 {\scriptsize(21.9)} & .236 & 4.47$\times$ & \textbf{.236} & 0.25 & \textbf{6.64$\times$\,(+49\%)} & \textbf{.234} {\scriptsize(.25)} & 0.03 & \textbf{6.79$\times$\,(+52\%)} & .232 / 7.80$\times$ \\
MBPP     & $K = 8$  & .234 {\scriptsize(21.9)} & .236 & 4.47$\times$ & \textbf{.236} & 1.41 & \textbf{7.69$\times$\,(+72\%)} & \textbf{.236} {\scriptsize(.25)} & 0.56 & \textbf{8.93$\times$\,(+100\%)} & .228 / 12.50$\times$ \\
\bottomrule
\end{tabular}}
\end{center}
\vspace{-5pt}
\end{table}

Table~\ref{tab:wake} highlights three findings. First, at $\tau = 0$, Wake verification
accelerates decoding beyond the $K = T$ block baseline while preserving greedy-exact
$T$-pass fidelity: on Ouro-2.6B, $K = 2$ Exact Wake matches $K = 4$ block accuracy to
four decimal places across tasks (with committed-token audit discrepancies of $0.007\%$
to $0.09\%$, below the $0.25\%$ to $0.33\%$ sequential \texttt{bf16} floor) while
adding \textbf{+22\% to +28\%} incremental throughput over $K = 4$ (\textbf{5.20$\times$}
on GSM8K, outpacing even non-rollback lossy $K = 3$ at $5.18\times$, and
\textbf{5.23$\times$} on MATH500). On Huginn-3.5B ($T = 32$), because late-entry
predictions already agree on $> 98\%$ of positions, rollbacks stay sparse ($0.25$ to
$2.22$ per $100$ tokens), lifting $\tau = 0$ greedy-exact speedups to \textbf{4.85 to
6.64$\times$} (\textbf{+49\% to +66\%} over $K = 32$ block) at $K = 16$ ($K = 12$
reaches $5.87\times$, $+75\%$) and \textbf{6.09 to 7.69$\times$} (\textbf{+72\% to
+122\%} over $K = 32$ block) at $K = 8$, completely eliminating four-stage $K = 8$
GSM8K lag ($.353 \to .392$).
Second, margin gating ($\tau \in [0.15, 0.30]$) filters benign near-tie flips ($\Delta z
\le \tau$), cutting rollbacks $2\times$ to $16\times$ and sustaining full accuracy at
\textbf{5.52 to 5.83$\times$} on Ouro GSM8K and MATH500 ($\mathbf{5.77\times}$ at
$\tau = 0.30$ and $\mathbf{6.02\times}$ at $\tau = 0.60$ on GSM8K, $p \ge .30$) and at
\textbf{5.10 to 6.79$\times$} ($K = 16$) and \textbf{7.15 to 8.93$\times$} ($K = 8$) on
Huginn. Third, testing our second parameter-free
$4/3/2$-gram context-suffix drafter alongside trained mask placeholders confirms the
drafter-verifier factorization analyzed in Section~\ref{sec:roofline}
(Figure~\ref{fig:wake}, violet curves): although unconditional short-ngram override slightly lowers raw block
acceptance on math and code relative to trained 48-layer placeholder lookahead ($3.93$
vs.\ $5.02$ tokens/step on Ouro GSM8K; $4.49$ vs.\ $5.34$ on Huginn MBPP; while matching
or slightly outpacing it on Ouro IFEval at $2.68\times$ vs.\ $2.66\times$), the
multiplicative speedup gain from $K < T$ Wake verification holds or strengthens across
both drafters ($1.37\times$ at $\tau = 0$ and $1.48\times$ at $\tau = 0.25$ vs.\
$1.26\times$ and $1.40\times$ on Ouro GSM8K; $1.89\times$ vs.\ $1.72\times$ on Huginn
MBPP; $2.25\times$ vs.\ $2.22\times$ on Huginn IFEval).

\subsection{Intra-Model Recurrence Depth Scaling ($T \in \{8, 16, 32, 64\}$)}
\label{sec:tscale}

Comparing Ouro-2.6B ($T = 4$) and Huginn-3.5B ($T = 32$) leaves recurrence depth $T$
partially confounded with model family and readout architecture. Because Huginn's
recurrent Core shares weights across iterations and generalizes to arbitrary test-time
loop counts \citep{geiping2025scaling}, we isolate $T$ directly by evaluating the
\emph{exact same} Huginn-3.5B checkpoint across four recurrence depths $T \in \{8, 16,
32, 64\}$ ($36, 68, 132,$ and $260$ effective Transformer layers per token) on full
GSM8K ($n = 1{,}319$) inside the identical CUDA-graphed engine.

\begin{figure}[!t]
\begin{center}
\includegraphics[width=0.95\textwidth]{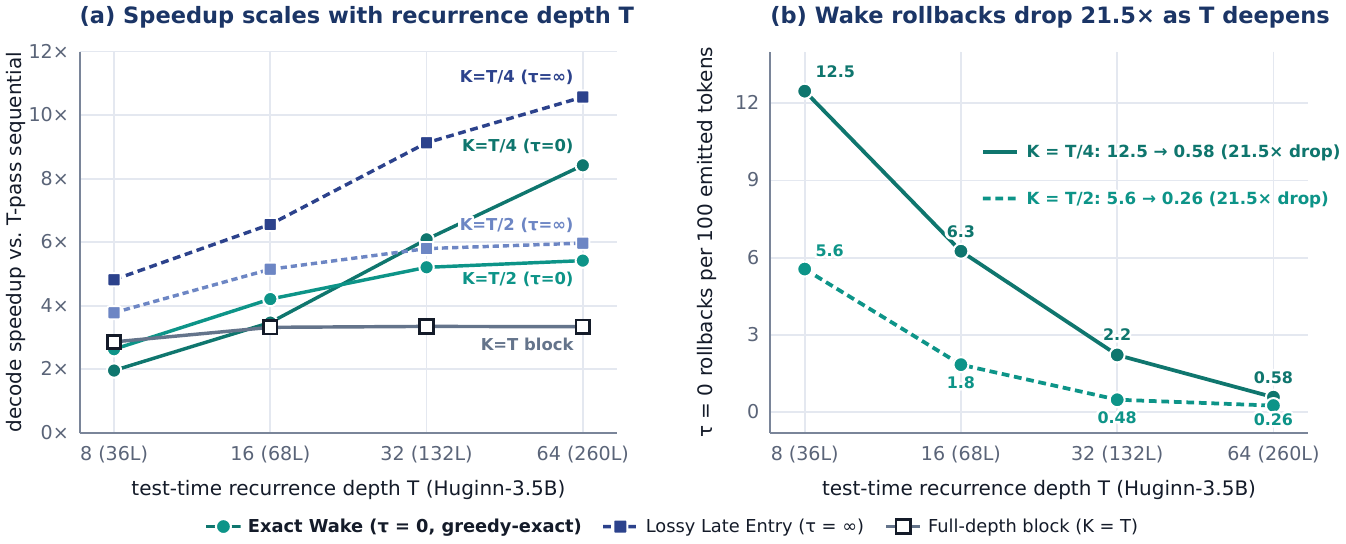}
\end{center}
\vspace{-6pt}
\caption{\textbf{Intra-model recurrence depth scaling on Huginn-3.5B across $T \in \{8,
16, 32, 64\}$} ($36 \to 260$ effective Transformer layers/token, identical checkpoint
and engine, full GSM8K $n = 1{,}319$). \textbf{(a)~Decode speedup vs.\ $T$:} while
full-depth $K = T$ block speedup plateaus near $3.3\times$, $\tau = 0$ greedy-exact Wake
scales from $1.96\times$ ($T = 8$) to \textbf{8.42$\times$} ($T = 64$, \textbf{+153\%}
over $K = 64$ block) and $\tau = \infty$ late entry reaches \textbf{10.57$\times$}.
\textbf{(b)~Wake rollback collapse:} $\tau = 0$ rollbacks per $100$ emitted tokens fall
\textbf{21.5$\times$} across $T = 8 \to 64$ at both $K = T/4$ ($12.5 \to 0.58$) and
$K = T/2$ ($5.6 \to 0.26$).}
\label{fig:tscale}
\vspace{-4pt}
\end{figure}

\begin{table}[!t]
\caption{\textbf{Intra-model recurrence depth scaling on Huginn-3.5B across $T \in \{8,
16, 32, 64\}$ on full GSM8K} ($n = 1{,}319, B = 8$, identical checkpoint and engine).
``rb'' reports $\tau = 0$ rollbacks per $100$ tokens ($\dagger$ marks paired sign-test
$p < .05$ vs.\ $T$-pass sequential decoding at the same $T$).}
\label{tab:tscale}
\begin{center}
\small
\setlength{\tabcolsep}{3.8pt}
\shrinktofit{%
\begin{tabular}{lccrcccrccc}
\toprule
& \multicolumn{2}{c}{Seq.\ $T$-Pass} & $K\!=\!T$ & \multicolumn{3}{c}{\textbf{Half Depth ($K = T / 2$)}} & \multicolumn{4}{c}{\textbf{Quarter Depth ($K = T / 4$)}} \\
\cmidrule(lr){2-3}\cmidrule(lr){4-4}\cmidrule(lr){5-7}\cmidrule(lr){8-11}
Depth $T$ (Eff.\ Layers) & Acc. & tok/s & Spd. & $\tau\!=\!0$ Acc.\ / rb & $\tau\!=\!0$ Spd. & $\tau\!=\!\infty$ Acc.\ / Spd. & $\tau\!=\!0$ Acc. & rb & $\tau\!=\!0$ Spd. & $\tau\!=\!\infty$ Acc.\ / Spd. \\
\midrule
$T = 8$ \ ($36$ layers)   & .267 & 78.6 & 2.86$\times$ & \textbf{.265} / 5.56 & 2.63$\times$          & .167$^\dagger$ / 3.78$\times$ & \textbf{.262} & 12.46 & 1.96$\times$          & .030$^\dagger$ / 4.82$\times$ \\
$T = 16$ ($68$ layers)    & .382 & 42.1 & 3.32$\times$ & \textbf{.379} / 1.85 & \textbf{4.21$\times$} & .361\phantom{$^\dagger$} / 5.15$\times$ & \textbf{.375} & \phantom{0}6.25 & \textbf{3.47$\times$} & .146$^\dagger$ / 6.56$\times$ \\
$T = 32$ ($132$ layers)   & .391 & 21.9 & 3.35$\times$ & \textbf{.396} / 0.48 & \textbf{5.21$\times$} & .399\phantom{$^\dagger$} / 5.80$\times$ & \textbf{.392} & \phantom{0}2.22 & \textbf{6.09$\times$} & .353$^\dagger$ / 9.13$\times$ \\
$T = 64$ ($260$ layers)   & .397 & 11.1 & 3.34$\times$ & \textbf{.397} / 0.26 & \textbf{5.42$\times$} & .394\phantom{$^\dagger$} / 5.97$\times$ & \textbf{.395} & \phantom{0}0.58 & \textbf{8.42$\times$} & \textbf{.392} / \textbf{10.57$\times$} \\
\bottomrule
\end{tabular}}
\end{center}
\vspace{-5pt}
\end{table}

Figure~\ref{fig:tscale} and Table~\ref{tab:tscale} reveal a clean depth-scaling law. As
test-time recurrence doubles from $T = 8 \to 16 \to 32 \to 64$, sequential GSM8K
accuracy rises monotonically ($.267 \to .382 \to .391 \to .397$) while sequential
throughput halves at each step ($78.6 \to 11.1$~tokens/s). Across all four depths,
full-depth $K = T$ block speedup saturates flatly between $2.86\times$ and $3.35\times$
because block acceptance ($1 + \mathbb{E}[c] \approx 3.9$) is bounded by single-step
predictability rather than $T$. By contrast, late entry scales strongly with $T$:
consistent with progressive late-pass stabilization as recurrence deepens
\citep{geiping2025scaling}, $\tau = 0$ rollbacks per $100$ emitted tokens fall
\textbf{21.5$\times$} across $T = 8 \to 64$ at both $K = T/2$ ($5.56 \to 1.85 \to 0.48
\to 0.26$) and $K = T/4$ ($12.46 \to 6.25 \to 2.22 \to 0.58$;
Figure~\ref{fig:tscale}b). At shallow $T = 8$, $K = T/4 = 2$ Core passes is too shallow
for unverified readout and $\tau = \infty$ drops to $.030$; yet $\tau = 0$ Wake
\emph{catches every deviation and restores $.262$ accuracy} (\textbf{+23.2 points}),
throttling speed to $1.96\times$ when rollback cost exceeds the $T/K - 1$ threshold
derived next in Section~\ref{sec:roofline}. Once $T \ge 16$, $\tau = 0$ greedy-exact
Wake surpasses the $K = T$ ceiling and climbs to \textbf{5.42$\times$} ($K = 32$) and
\textbf{8.42$\times$} ($K = 16$, $+153\%$ over $K = 64$ block, with $\tau = 0.25$ at
\textbf{9.17$\times$}) at $T = 64$, where even non-rollback $\tau = \infty$ ($K = 16$)
retains \textbf{.392} accuracy ($p = .45$) at \textbf{10.57$\times$} speedup.

\subsection{When Does Late Entry Remain Fast? Roofline and Multi-Request Scaling}
\label{sec:roofline}

\begin{figure}[!t]
\begin{center}
\includegraphics[width=0.96\textwidth]{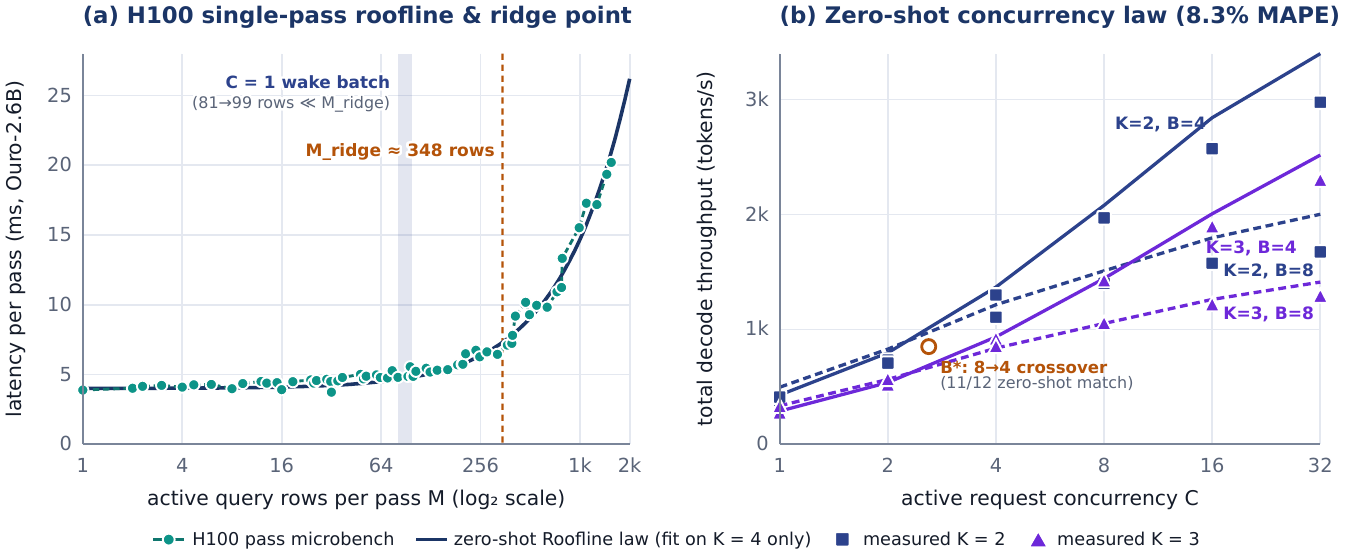}
\end{center}
\vspace{-6pt}
\caption{\textbf{Physical H100 Roofline and zero-shot multi-request throughput law on
Ouro-2.6B ($T = 4$).} \textbf{(a)~Single-pass Roofline:} at batch size $C = 1$,
appending wake catch-up rows ($81 \to 99$ rows) sits deep inside the HBM
weight-streaming plateau well below the hardware ridge point $M_{\mathrm{ridge}}(1)
\approx 348$ rows/pass. \textbf{(b)~Zero-shot concurrency prediction:} calibrating
Equation~\eqref{eq:roofline} \emph{strictly on full-depth $K = 4$ modes} predicts
unseen $K \in \{2, 3\}$ continuous-batching throughput across $C \in [1, 32]$ with
\textbf{8.3\% MAPE} and selects the optimal block size $B^*(C) \in \{4, 8\}$ on
\textbf{11/12 (91.7\%)} operating points.}
\label{fig:roofline}
\vspace{-4pt}
\end{figure}

Finally, to explain from first principles why piggybacked wake rows stay nearly free at
batch~$1$ and how throughput scales as concurrent serving load $C \in [1, 32]$ saturates
GPU compute (full continuous-batching sweeps in Appendix~\ref{app:concurrency}), we
model single-pass GPU latency for $M$ active query rows across $C$ concurrent requests
via an $L_p$ Roofline law:
\begin{equation}
  t_{\mathrm{pass}}(M, C) = \Big(\big(t_0 + \alpha_{\mathrm{kv}}\,C\big)^p
    + \big(\beta_{\mathrm{tc}}\,M\big)^p\Big)^{1/p},
  \qquad
  \Theta(K, B, C, \tau) = \frac{C \cdot \bar{c}_{\mathrm{net}}(K, B, \tau)}
    {\sum_{j=1}^{K} t_{\mathrm{pass}}\!\big(M_j(K, B, C),\, C\big)},
  \label{eq:roofline}
\end{equation}
where $t_0$ is single-pass HBM weight-streaming latency, $\alpha_{\mathrm{kv}} C$ is
KV-cache read overhead, $\beta_{\mathrm{tc}} M$ is Tensor-Core compute time, and
$\bar{c}_{\mathrm{net}}(K, B, \tau)$ is net committed tokens per step after Wake
rollbacks. Calibrating $(t_0 = 3.95\,\mathrm{ms},\, \alpha_{\mathrm{kv}} =
0.05\,\mathrm{ms},\, \beta_{\mathrm{tc}} = 0.0115\,\mathrm{ms/row},\, p = 1.15)$
\emph{exclusively on full-depth $K = 4$ modes} ($n = 18$ points, zero $K < 4$ latency
fits) locates the $C = 1$ hardware ridge point at $M_{\mathrm{ridge}}(1) =
(t_0 + \alpha_{\mathrm{kv}}) / \beta_{\mathrm{tc}} \approx 348$ rows/pass
(Figure~\ref{fig:roofline}a). Equation~\eqref{eq:roofline} ties together three empirical
behaviors observed across Sections~\ref{sec:frontier} to~\ref{sec:tscale}: (1)~at
$C = 1$, $M_j \le 99 \ll 348$, so wake completion and Wake verification execute inside
the weight-streaming floor ($T_{\mathrm{step}} \approx K t_0$), making $\tau = 0$
greedy-exact Wake strictly faster than $K = T$ block decoding whenever rollback
overhead stays below $T / K - 1$ (explaining both the $+22\%$ to $+153\%$ gains at
$T \ge 4$ and the graceful throttle at $T = 8, K = 2$); (2)~candidate drafting
($\bar{c}_{\mathrm{net}}$ in the numerator) decouples from pass-budget verification
($\sum_{j=1}^{K} t_{\mathrm{pass}}$ in the denominator), explaining why $K < T$ Wake
multipliers transfer across trained mask-placeholder and parameter-free n-gram drafters
(Section~\ref{sec:exp_wake}); and (3)~across $24$ unseen $K \in \{2, 3\}$ multi-request
points ($C \in \{1, \dots, 32\}$; Figure~\ref{fig:roofline}b),
Equation~\eqref{eq:roofline} predicts throughput zero-shot at \textbf{8.3\% MAPE} and
identifies the optimal block-size crossover $B^*(C)$ on \textbf{11/12 (91.7\%)}
operating points, guiding dynamic $B = 8 \to 4 \to 2$ profile selection up to $C = 32$
(\textbf{2.31$\times$} decode, \textbf{1.67$\times$} end-to-end;
Appendix~\ref{app:concurrency}).

\section{Related Work}
\label{sec:related}

Looped and recursive language models iterate a weight-tied layer stack to scale latent
reasoning depth at a compact parameter footprint
\citep{dehghani2019universal,bae2025relaxed,bae2025mor,geiping2025scaling,ouro2025}.
In standard non-recurrent Transformers, SkipDecode \citep{delcorro2023skipdecode}
skips bottom layers in favor of upper layers under a monotonically decreasing per-step
layer budget so KV caches never require recomputation, while in shared-weight recursive
models Relaxed Recursive Transformers \citep{bae2025relaxed} introduce cross-request
depth-wise batching and Mixture-of-Recursions \citep{bae2025mor} learns token-level
routing across initial passes. Our analysis addresses a distinct regime: because a
looped model shares weights across all passes, every token can enter uniformly at the
final $K$ passes without decaying position budgets, while within-sequence pipelined
cache completion reconstructs exact $1 \to T$ context caches in the background wake.

To accelerate autoregressive decoding, speculative and block-decoding methods verify
multiple drafted tokens in parallel using auxiliary drafters, n-gram suffix lookups,
or single-pass mask placeholders
\citep{leviathan2023fast,chen2023accelerating,cai2024medusa,fu2024lookahead,gloeckle2024better,liu2025tidar,jin2026mars}.
For looped language models specifically, WaveFront \citep{ha2026wavefront}, LoopSpec
\citep{cho2026loopspec}, DAS \citep{li2026das}, and LoopAhead \citep{liu2026loopahead}
accelerate \emph{drafting} via shallow initial passes: DAS and LoopAhead verify
candidates synchronously through a full $T$-pass target call, whereas WaveFront and
LoopSpec pipeline single-pass shallow guesses diagonally into deeper passes.
Crucially, because those methods only speed up proposal generation while still routing
every accepted token through shallow-to-deep passes on the verification path, full
$T$-pass verification bounds their speedup as recurrence deepens. Late entry paired with
Wake verification is complementary and structurally distinct along three axes:
(1)~our optimistic frontier executes the \emph{final} $K < T$ passes over a deep context
cache rather than initial passes over shallow caches, cutting mismatch and rollback
rates by an order of magnitude ($0.25$ to $2.2$ rollbacks per $100$ tokens at $T \ge 16$);
(2)~in-pass block verification filters $(1 + B)^2$ candidate branches at depth $K$ so
only the $\le 1 + B$ committed tokens enter the $1 \to T$ wake, saving $T - K$ passes on
all rejected branches; and (3)~a single static CUDA graph sweeps $\tau \in [0, \infty]$
from greedy-exact ($\tau = 0$) to margin-gated ($\tau > 0$) and lossy ($\tau = \infty$)
decoding while factorizing orthogonally from the choice of candidate drafter.

Finally, adaptive early-exit methods halt computation after initial layers or passes
when confidence is high \citep{schuster2022calm}, while loop-contrastive methods
contrast early and late passes to sharpen reasoning accuracy
\citep{yu2026loopcd,liu2026loopcd}. We show that halting early without cache
completion starves a looped language model of coherent pass-aligned $(K, V)$ context
depth, whereas optimistic late entry paired with Wake verification preserves full-depth
context and exactness while advancing at $K < T$ serial passes per step.

\section{Limitations and Conclusion}
\label{sec:conclusion}

Our study has three main limitations. First, as formalized by our Roofline law in
Section~\ref{sec:roofline}, single-request speedups exploit HBM weight-streaming
headroom on an H100 GPU ($M_j \le 99 \ll M_{\mathrm{ridge}}(1) \approx 348$ rows/pass).
Under multi-request continuous batching on Ouro-2.6B (Figure~\ref{fig:roofline}b;
Appendix~\ref{app:concurrency}), speedup compression across $C = 1 \to 32$ stems
primarily from $(1 + B)^2$ block expansion rather than late entry itself: full-depth
$B = 8$ block decoding drops from $3.90\times$ to $0.77\times$, whereas late entry's
incremental gain over same-$B$ block decoding holds steady ($1.65\times \to 1.45\times$
at $B = 8$ and $1.84\times \to 1.48\times$ at $B = 4$ for $K = 2$) because skipping
$T - K$ passes on rejected draft branches cuts total arithmetic work per step ($198$
vs.\ $324$ row-passes at $B = 8$). Guided by Equation~\eqref{eq:roofline} ($8.3\%$
zero-shot MAPE, $11/12$ optimal $B^*(C)$ matches), shrinking $B = 8 \to 4 \to 2$ with
load retains $2.31\times$ decode ($1.67\times$ end-to-end including online prefills) at
$C = 32$ for $K = 2$ ($1.71\times$ and $1.47\times$ for $K = 3$).
Second, while Wake verification unifies greedy-exact ($\tau = 0$), margin-gated
($\tau > 0$), and lossy ($\tau = \infty$) decoding inside one engine, Proposition~1
targets greedy decoding (extending Wake rollback to temperature sampling via speculative
rejection sampling \citep{leviathan2023fast,chen2023accelerating} is left to future
work), and $\tau = 0$ speedups depend on optimistic agreement: at shallow recurrence
($T = 8, K = 2$), frequent rollbacks ($12.5$ per $100$ tokens) preserve full accuracy
($+23.2$ points over $\tau = \infty$) at the cost of falling below $K = T$ block speed
($1.96\times$ vs.\ $2.86\times$), whereas for $T \ge 16$ ($0.26$ to $2.2$ rollbacks per
$100$ tokens) $\tau = 0$ scales up to $8.42\times$. Third, unlocking late entry on
checkpoints not pretrained with pass-skipping requires a one-time Stage-2 SFT pass
($7{,}395$ steps, $\approx 1.9$B tokens on one $8\times$H100 node, comparable to
training lightweight speculative heads such as Medusa or LoopAhead while adding zero
inference parameters; Appendix~\ref{app:arms}), motivating native late-entry exposure
during pretraining and per-token adaptive pass scheduling in future work.

In conclusion, saving passes from the two ends of a looped language model is
fundamentally asymmetric: cached context depth substitutes for token-local early
passes across Ouro-1.4B, Ouro-2.6B ($T = 4$), and middle-recurrent Huginn-3.5B
($T \in \{8, 16, 32, 64\}$). Pairing late-pass block verification with pipelined cache
completion and zero-extra-pass Wake verification turns this asymmetry into a unified
$\tau \in [0, \infty]$ family that accelerates both greedy-exact $T$-pass decoding (up
to $5.2\times$ at $T = 4$ and $8.4\times$ at $T = 64$) and controlled lossy decoding
(up to $6.8\times$ and $12.5\times$). Engine code, training
scripts, and checkpoints are available at
\url{https://github.com/Xalp/LateComer} and
\url{https://huggingface.co/collections/Xalphinions/late-entry} (project page:
\url{https://xalp.net/late-entry/}).

\bibliographystyle{plainnat}
\bibliography{references}

\appendix

\section{Fine-Tuning Recipe and Diagnostic Probe Protocol}
\label{app:recipe}

\paragraph{Two-stage fine-tuning.}
We initialize Ouro-2.6B (and Ouro-1.4B for scale replication) from its released
pretrained checkpoint \citep{ouro2025} and fine-tune on Dolci-Instruct-SFT
\citep{olmo2025olmo3} in two
one-epoch stages. Stage~1 performs standard full-depth ($T = 4$) instruction SFT at
sequence length $1{,}024$. Stage~2 runs one epoch of MARS-style mask-placeholder
block fine-tuning \citep{jin2026mars} using the block structure of
Section~\ref{sec:blockverify}: blocks of $16$ token positions with mask-placeholder
branches, learning rate $10^{-5}$ with $3\%$ linear warmup, global batch size $256$ in
\texttt{bf16}, and sequence length $1{,}024$ ($7{,}395$ optimization steps, $\approx
1.9$B tokens per stage, adding zero extra parameters at inference). During Stage~2, half of the blocks run all four
passes starting from pass~$1$, while the other half enter directly at pass
$T - K + 1$ ($K \sim \mathrm{Uniform}\{1, 2, 3\}$) from raw token embeddings. Our
matched early-exit control branches from the exact same Stage-1 checkpoint with
identical data order, seed, and hyperparameters, reversing only the reduced-budget
subset so tokens execute the first $K$ passes instead of the last $K$.

\paragraph{Ouro stick-breaking readout and teacher-forced probe protocol.}
When Ouro executes a contiguous pass range $\{a, \dots, b\} \subseteq \{1, \dots, T\}$,
its learned stick-breaking exit gate emits a scalar gate probability
$\lambda_i^{(k)} \in (0, 1)$ at each executed pass $k$, assigning weight
$w_i^{(k)} = \lambda_i^{(k)} \prod_{j=a}^{k-1} (1 - \lambda_i^{(j)})$ to passes $k < b$
and all remaining mass $w_i^{(b)} = \prod_{j=a}^{b-1} (1 - \lambda_i^{(j)})$ to pass $b$
(so $\sum_{k=a}^{b} w_i^{(k)} = 1$ without requiring skipped passes):
\begin{equation}
  p_{a:b}(x_{i+1} \mid x_{\le i}) = \mathrm{softmax}\!\left(
    W_{\mathrm{out}} \sum_{k = a}^{b} w_i^{(k)}\, \tilde{h}_i^{(k)}
  \right).
  \label{eq:readout}
\end{equation}
To measure single-step agreement without trajectory drift, we first decode full test
prompts greedily at full depth ($T = 4$, batch size~$1$) to obtain fixed reference
completions ($70{,}101$ GSM8K token positions on Ouro-2.6B and $78{,}693$ on
Ouro-1.4B, with the identical procedure on IFEval). We then replay each reference
sequence under teacher forcing using Equation~\eqref{eq:readout} and record the fraction
of positions whose greedy prediction matches the full-depth $\arg\max$.

\section{Training-Stage Ablation}
\label{app:arms}

Table~\ref{tab:arms} disentangles the effects of instruction tuning, block drafting,
and late-entry exposure on Ouro-2.6B. Neither Stage-1 SFT alone ($.699$ probe
agreement) nor Stage-2 block training without late-entry exposure ($.738$) enables
late entry. Explicit late-entry exposure during Stage~2 lifts probe agreement to
\textbf{.962} and prevents the severe $K = 2$ decoding collapse seen in the unexposed
block model ($.636$ vs.\ \textbf{.839} on GSM8K; $.529$ vs.\ \textbf{.699} on
IFEval). It also preserves full-depth GSM8K accuracy ($.873$ vs.\ $.880$ after
Stage~1, whereas the unadapted base model scores $.729$ and generates $243$ mean
tokens compared with $114$ after SFT).

\begin{table}[!t]
\caption{\textbf{Training-stage ablation on Ouro-2.6B} across full GSM8K ($n = 1{,}319$)
and strict IFEval ($n = 541$). The probe column reports teacher-forced agreement at
$K = 1, D = 4$; the rightmost column reports end-to-end block decoding accuracy at
$K = 2, B = 8$.}
\label{tab:arms}
\begin{center}
\small
\setlength{\tabcolsep}{3.0pt}
\begin{tabular}{lccc}
\toprule
Training configuration & Full-depth acc.\ (GSM8K / IFEval) & Late-entry probe & Late-entry decode ($K = 2$) \\
\midrule
None (released base model)            & .729 / .442 & .751 & n/a (no drafters) \\
Stage~1 only (full-depth SFT)         & .880 / .728 & .699 & n/a (no drafters) \\
Stage~1 + Stage~2, late entry off     & .820 / .708 & .738 & .636 / .529 \\
Stage~1 + Stage~2 (our full recipe)   & .873 / .732 & \textbf{.962} & \textbf{.839} / \textbf{.699} \\
\bottomrule
\end{tabular}
\end{center}
\end{table}

\section{Extended Entry-State and Attention-Geometry Diagnostics}
\label{app:entry}

Table~\ref{tab:entryfull} expands the off-the-shelf diagnostics from
Table~\ref{tab:twosizes}. Naively rescaling normalized embeddings ($0.76$ norm) to
match pass-4 state norms ($37$ to $51$) worsens Ouro-2.6B agreement to $.218$.
Feeding the preceding token's deep hidden state alone causes severe next-token
repetition ($\le .039$, although combining it with a linear embedding projection
reaches $.949$). Using unembedding vectors likewise repeats the current token ($.01$
to $.04$), while soft probability-weighted embeddings match hard embeddings on over
$99\%$ of confident positions. Even when we train two-layer MLP entry adapters while
keeping Ouro weights frozen, agreement trails simple $K$-pass truncation by $1.1$,
$3.4$, and $2.9$ points at $K = 1, 2, 3$ on Ouro-2.6B (and fails to transfer to
IFEval on Ouro-1.4B). On IFEval, closed-form linear maps recover $46\%$ to $51\%$ of
the oracle gap ($68\%$ to $72\%$ when combined with the predecessor state).

\paragraph{Attention-entropy, sink-mass, and distance-resolved KV interventions.}
Complementing Figure~\ref{fig:mechconf} in Section~\ref{sec:decouple}, our \texttt{fp32}
attention-geometry probes across $30{,}736$ GSM8K and $53{,}184$ IFEval positions show
that on unadapted checkpoints (released base, Stage~1 only, and Stage~2 without
late-entry exposure), feeding $e(x_i)$ directly into pass~$4$ ($K = 1, D = 4$) produces
diffuse attention ($+0.79$ to $+1.00$~nats higher head-mean entropy than a true pass-4
query and $-0.10$ to $-0.16$ less attention-sink mass on position~$0$), landing
off-manifold with linear CKA $.273$ to $.365$ against true pass-4 hidden states. Stage-2
late-entry exposure matches true pass-4 attention entropy within $+0.07$~nats on GSM8K
($+0.13$~nats on IFEval), restores sink mass within $-0.022$, and reaches linear CKA
$.961$ (mean cosine similarity $.927$ to $.930$). Distance-resolved prefix-depth
interventions further show that while adjacent tokens carry the largest per-token
marginal effect ($+1.6$ agreement points at distance~$1$), deep caches more than $64$
tokens back still account for $14\%$ of the total context-depth gain on GSM8K and
\textbf{53\%} ($+6.9$ points) on longer IFEval generations.

\begin{table}[!t]
\caption{\textbf{Extended pass-4 entry-state diagnostics on off-the-shelf Ouro-1.4B
and Ouro-2.6B} (GSM8K teacher-forced agreement with full depth; top five rows run
pass~$4$ only over a four-pass context cache).}
\label{tab:entryfull}
\begin{center}
\small
\begin{tabular}{lcc}
\toprule
Input state supplied to pass~4 ($K = 1$ over four-pass context) & Released Ouro-1.4B & Released Ouro-2.6B \\
\midrule
Raw token embedding \emph{(unadapted late entry)}         & .552 & .751 \\
Normalized and magnitude-rescaled embedding               & .245 & .218 \\
Closed-form linear projection of the embedding            & .816 & .919 \\
Preceding token's deep hidden state alone                 & .030 & .039 \\
True pass-3 hidden state \emph{(diagnostic oracle ceiling)} & .991 & .992 \\
\midrule
\emph{Reference:} 1-pass early exit (pass~1 only)         & .938 & .949 \\
\emph{Reference:} 2-pass early exit (passes~1 and~2 only) & .981 & .985 \\
\bottomrule
\end{tabular}
\end{center}
\end{table}

\section{Complete $(K, B)$ Frontier Grid and Verification Threshold Sweep}
\label{app:fullgrid}

Table~\ref{tab:frontierfull} reports the complete grid of pass budgets
$K \in \{1, 2, 3, 4\}$ and block sizes $B \in \{4, 8\}$ on GSM8K and IFEval. Using
$B = 8$ improves throughput by $12\%$ to $22\%$ over $B = 4$ on GSM8K while achieving
similar throughput and slightly higher accuracy on IFEval. To test whether we could
obtain similar speedups without skipping passes, Table~\ref{tab:thsweep} keeps full
four-pass verification ($K = 4, B = 8$) and relaxes acceptance so any candidate whose
full-depth probability exceeds a threshold $\theta \in \{.05, .10, .20, .30, .40\}$
is accepted. Every threshold $\theta$ is strictly Pareto-dominated on both benchmarks
by strict greedy ($\arg\max$) verification under late entry at $K = 3$ and $K = 2$.

\begin{table}[!t]
\caption{\textbf{Complete $(K, B)$ accuracy-throughput grid on full GSM8K and IFEval
test sets} (batch size~$1$, greedy decoding, single H100 GPU).}
\label{tab:frontierfull}
\begin{center}
\small
\begin{tabular}{llcccccc}
\toprule
& & \multicolumn{3}{c}{\textbf{GSM8K} ($n = 1{,}319$)} & \multicolumn{3}{c}{\textbf{IFEval} ($n = 541$)} \\
\cmidrule(lr){3-5}\cmidrule(lr){6-8}
Setting & $B$ & tok/s & Speedup & Acc. & tok/s & Speedup & Acc. \\
\midrule
Full-depth sequential & n/a & 63.4  & 1.00$\times$ & .873 & 63.1  & 1.00$\times$ & .732 \\
\midrule
$K = 4$ \ (\loopmark{4}) & 4 & 210.5 & 3.32$\times$ & .870 & 137.4 & 2.18$\times$ & .719 \\
                         & 8 & 256.6 & 4.05$\times$ & .867 & 136.7 & 2.16$\times$ & .739 \\
$K = 3$ \ (\loopmark{3}) & 4 & 269.7 & 4.26$\times$ & .862 & 175.0 & 2.77$\times$ & .728 \\
                         & 8 & 322.5 & 5.09$\times$ & .851 & 178.1 & 2.82$\times$ & .734 \\
$K = 2$ \ (\loopmark{2}) & 4 & 380.4 & 6.00$\times$ & .846 & 247.2 & 3.91$\times$ & .697 \\
                         & 8 & 426.7 & 6.73$\times$ & .839 & 238.6 & 3.78$\times$ & .699 \\
$K = 1$ \ (\loopmark{1}) & 4 & 539.1 & 8.51$\times$ & .637 & 385.3 & 6.10$\times$ & .625 \\
                         & 8 & 544.0 & 8.59$\times$ & .629 & 385.5 & 6.11$\times$ & .645 \\
\bottomrule
\end{tabular}
\end{center}
\end{table}

\begin{table}[!t]
\caption{\textbf{Relaxed probability-threshold verification at full depth
($K = 4, B = 8$) versus strict greedy ($\arg\max$) late entry.} Accepting proposals
whose full-depth probability exceeds $\theta$ yields smaller speedups and lower
accuracy than reducing $K$ under strict $\arg\max$ verification.}
\label{tab:thsweep}
\begin{center}
\small
\begin{tabular}{lcccc}
\toprule
& \multicolumn{2}{c}{\textbf{GSM8K}} & \multicolumn{2}{c}{\textbf{IFEval}} \\
\cmidrule(lr){2-3}\cmidrule(lr){4-5}
Verification rule & Speedup & Acc. & Speedup & Acc. \\
\midrule
Strict $\arg\max$, $K = 4$      & 4.06$\times$ & .867 & 2.17$\times$ & .739 \\
Relaxed $\theta = .40$, $K = 4$ & 4.14$\times$ & .867 & 2.19$\times$ & .726 \\
Relaxed $\theta = .30$, $K = 4$ & 4.30$\times$ & .861 & 2.26$\times$ & .730 \\
Relaxed $\theta = .20$, $K = 4$ & 4.40$\times$ & .841 & 2.36$\times$ & .719 \\
Relaxed $\theta = .10$, $K = 4$ & 4.49$\times$ & .834 & 2.56$\times$ & .734 \\
Relaxed $\theta = .05$, $K = 4$ & 4.53$\times$ & .823 & 2.69$\times$ & .728 \\
\midrule
Strict $\arg\max$, $K = 3$      & \textbf{5.09$\times$} & \textbf{.851} & \textbf{2.82$\times$} & \textbf{.734} \\
Strict $\arg\max$, $K = 2$      & \textbf{6.73$\times$} & \textbf{.839} & \textbf{3.78$\times$} & .699 \\
\bottomrule
\end{tabular}
\end{center}
\end{table}

\section{Committed-Token Fidelity Audit}
\label{app:audit}

To verify exactness at $K = 4$ and quantify approximation at $K < 4$, we replay all
committed sequences through teacher-forced full-depth forward passes
(Table~\ref{tab:audit}). At $K = 4$, the token-level discrepancy rate ($0.01\%$ to
$0.41\%$, with median top-2 logit margins of $.03$ to $.07$) lies at or below the
sequential \texttt{bf16} kernel-order floor ($0.25\%$ to $0.64\%$), and all
$3{,}912/3{,}912$ dense structural checks match identically. Our early-exit block
decoder passes the same audit ($0.03\%$ discrepancy vs.\ a $0.23\%$ floor at $K = 2$
on GSM8K; $24/24$ dense checks). At $K = 3$ and $K = 2$ without Wake verification
($\tau = \infty$), token discrepancy rises moderately to $1.0\%$ to $5.9\%$; enabling
Exact Wake verification ($\tau = 0$, Section~\ref{sec:wake}) restores the exact $K = 4$
fidelity floor ($0.007\%$ on GSM8K, $0.09\%$ on MATH500, and $0.025\%$ on MBPP).

\begin{table}[!t]
\caption{\textbf{Committed-token fidelity audit at block size $B = 8$.} ``diff'' is
the percentage of committed positions whose token differs from a teacher-forced
full-depth $\arg\max$ replay; ``margin'' reports the median (and 90th percentile)
top-2 logit margin at differing positions.}
\label{tab:audit}
\begin{center}
\small
\setlength{\tabcolsep}{3.6pt}
\begin{tabular}{lcccccccc}
\toprule
& \multicolumn{2}{c}{Full-depth floor} & \multicolumn{2}{c}{\loopmark{4}\;$K = 4$} & \multicolumn{2}{c}{\loopmark{3}\;$K = 3$} & \multicolumn{2}{c}{\loopmark{2}\;$K = 2$} \\
\cmidrule(lr){2-3}\cmidrule(lr){4-5}\cmidrule(lr){6-7}\cmidrule(lr){8-9}
Benchmark & diff & margin & diff & margin & diff & margin & diff & margin \\
\midrule
GSM8K    & 0.27\% & .04 (.10) & 0.01\% & .06 (.11) & 1.12\% & .11 (.48) & 2.08\% & .22 (.87) \\
IFEval   & 0.64\% & .03 (.09) & 0.41\% & .03 (.09) & 3.17\% & .13 (.50) & 5.86\% & .23 (.77) \\
MATH500  & 0.25\% & .05 (.11) & 0.09\% & .05 (.12) & 1.16\% & .13 (.53) & 2.16\% & .23 (1.00) \\
MBPP     & 0.33\% & .04 (.10) & 0.02\% & .07 (.30) & 1.04\% & .10 (.39) & 1.87\% & .18 (.71) \\
BBH      & 0.27\% & .04 (.11) & 0.06\% & .04 (.12) & 1.33\% & .15 (.77) & 2.43\% & .33 (1.41) \\
MMLU-Pro & 0.32\% & .04 (.11) & 0.09\% & .05 (.12) & 1.41\% & .13 (.52) & 2.69\% & .25 (1.07) \\
\bottomrule
\end{tabular}
\end{center}
\end{table}

\section{Cache-Completion Ablations and Error Localization}
\label{app:catchup}

Table~\ref{tab:catchup} compares three cache-completion policies on GSM8K and IFEval.
Our lagged pipelined completion matches or slightly exceeds synchronous
before-next-step repair ($p = .28$ to $.77$), confirming that reading provisional KV
entries for one or two steps introduces zero measurable degradation. Conversely,
omitting cache completion altogether (keeping provisional embedding-derived KV states
permanently) drops GSM8K accuracy by \textbf{4.1 points} at $K = 2$ ($.839 \to
.798$).

\begin{table}[!t]
\caption{\textbf{Cache-completion policy ablation} on full GSM8K ($n = 1{,}319$) and
IFEval ($n = 541$). Full-depth sequential references are $.873$ and $.732$.}
\label{tab:catchup}
\begin{center}
\small
\begin{tabular}{llcccc}
\toprule
& & \multicolumn{2}{c}{\textbf{GSM8K}} & \multicolumn{2}{c}{\textbf{IFEval}} \\
\cmidrule(lr){3-4}\cmidrule(lr){5-6}
Cache-completion policy & Decoding mode & $K = 2$ & $K = 3$ & $K = 2$ & $K = 3$ \\
\midrule
Synchronous (repair before next step) & Sequential     & .852 & .864 & .684 & .717 \\
Pipelined (lagged across steps, ours) & Sequential     & .859 & .872 & .695 & .723 \\
Pipelined (lagged across steps, ours) & Block ($B = 8$) & \textbf{.839} & \textbf{.851} & \textbf{.699} & \textbf{.734} \\
Never (keep provisional cache)        & Block ($B = 8$) & .798 & .848 & .688 & .721 \\
\bottomrule
\end{tabular}
\end{center}
\end{table}

\begin{figure}[!t]
\begin{center}
\includegraphics[width=0.95\textwidth]{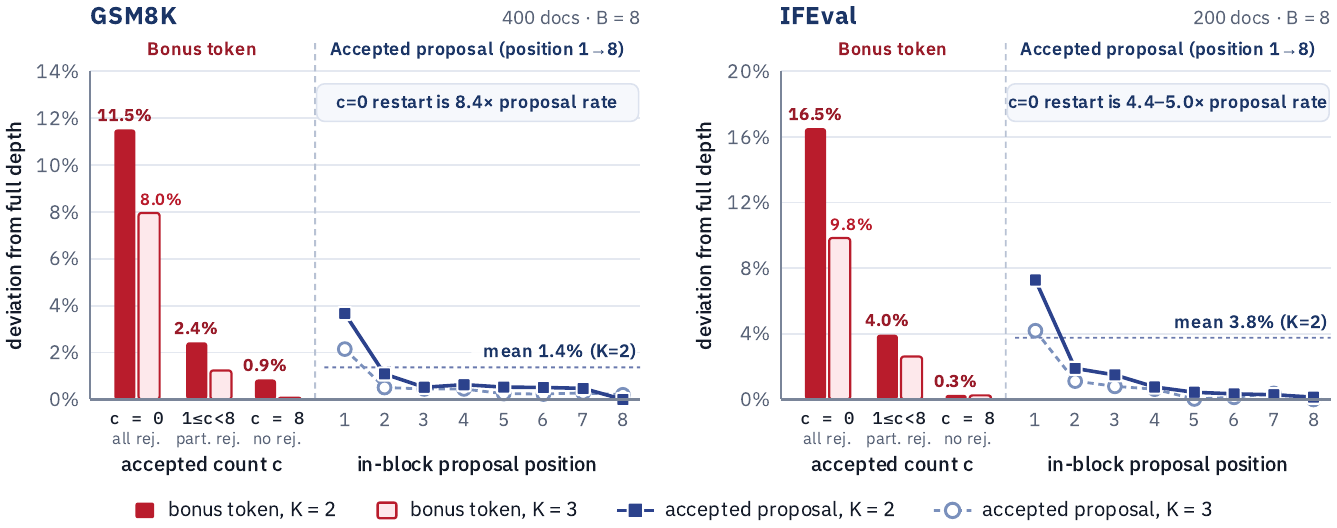}
\end{center}
\vspace{-4pt}
\caption{\textbf{Position-resolved deviation audit} at $B = 8$ and $K \in \{2, 3\}$
across $400$ GSM8K and $200$ IFEval examples. Left subpanels group bonus tokens by
accepted count $c$ ($c = 0$ means all $B = 8$ proposals were rejected); right
subpanels show deviation rates across accepted in-block positions $1 \to 8$.}
\label{fig:restarts}
\end{figure}

Where do the remaining token deviations at $K < 4$ occur within a block? Position-resolved
audits across $400$ GSM8K and $200$ IFEval items (Figure~\ref{fig:restarts}) show that
accepted proposal positions $1 \to 8$ average only $1.4\%$ deviation on GSM8K at
$K = 2$, while remaining discrepancies localize primarily to low-confidence
full-rejection steps where all $B$ proposals are rejected ($c = 0$). Exploiting this
localization, selectively recomputing only low-confidence ($< .5$) $c = 0$ steps ($19\%$
of IFEval steps) at full depth raises IFEval $K = 2$ accuracy from $.699$ to
\textbf{.723} ($p = .60$ vs.\ $.732$ full depth) at $196$~tokens/s, interpolating
smoothly between $K = 2$ ($239$~tokens/s) and $K = 3$ ($178$~tokens/s).

\section{Reproducibility, Adaptive Controls, and Full-Set Baseline Protocol}
\label{app:seqrepro}
\label{app:scorers}
\label{app:matched}

\paragraph{Sequential early-exit reproducibility and adaptive exit control.}
Independent sequential greedy evaluations of the trained early-exit checkpoint run
five days apart reproduce to four decimal places across all eight cells: GSM8K
$.3290, .6732, .7938, .8165$ and IFEval $.4510, .6026, .6858, .7116$ for
$K = 1, 2, 3, 4$. In addition to fixed-depth early exit in Table~\ref{tab:contest},
we also evaluate the trained early-exit model under a deliberately favored adaptive
exit control: each token chooses its exit pass via the model's learned exit gate with
thresholds calibrated directly on the test sets, while skipped passes are still
computed so subsequent tokens read an oracle full-depth KV cache. Even under these
favorable conditions, adaptive exit improves accuracy by less than $0.6$ points over
fixed-depth early exit at matched mean passes per token on both GSM8K and IFEval.

\paragraph{Gold-answer scorer validation.}
Running default \texttt{lm-evaluation-harness} extractors on dataset ground-truth
answers reveals two bugs: default MATH500 string matching rejects valid equivalent
expressions, and default MBPP code extraction strips line~1 of completions (causing
$499/500$ ground-truth programs to fail). Symbolic expression equivalence on MATH500
and preserved first-line extraction on MBPP restore $100\%$ accuracy on dataset
reference solutions.

\paragraph{Full-test-set baseline protocol.}
Table~\ref{tab:headtohead} evaluates the \emph{full} test set of every benchmark
($21{,}403$ prompts in total) with verified token-level prompt identity across
engines: every method decodes identical token-ID inputs at batch size~$1$ with greedy
decoding in \texttt{bf16} on a single NVIDIA H100 80GB GPU per job, and the two
largest benchmarks run as contiguous shards pooled exactly by total decode tokens
over total decode seconds. On each task, throughputs are normalized by the fastest
measured full-depth sequential decoder among ours, LoopSpec's, and the LoopAhead
reproduction's ($62.4$ to $67.2$~tokens/s; LoopSpec's engine is fastest on four
tasks, ours on MATH500 and MMLU-Pro). LoopSpec uses the fastest of four released
configurations selected on a full GSM8K sweep ($190.0$ vs.\ $184.8$, $173.9$, and
$119.5$~tokens/s). Because official LoopAhead code is unavailable, we report our
validated reproduction taking the fastest of three training arms per benchmark
($1\times$ and $4\times$ published training budget plus a no-drafter arm), evaluated
on the released checkpoint its method requires.

\section{Huginn-3.5B Architecture Adaptation and Extended Sweeps}
\label{app:huginn}

\paragraph{Prelude-Core-Coda late-entry formulation and training recipe.}
Unlike Ouro, which loops its entire Transformer stack and combines pass outputs via a
stick-breaking exit gate, Huginn-3.5B \citep[\texttt{huginn-0125};][]{geiping2025scaling}
wraps a recurrent $4$-layer Core block $\mathcal{F}_{\mathrm{core}}$ between a
$2$-layer non-recurrent Prelude $\mathcal{F}_{\mathrm{pre}}$ and a $2$-layer
non-recurrent Coda $\mathcal{F}_{\mathrm{coda}}$. For token $x_i$, Prelude maps the
scaled token embedding to a contextualized input anchor
$u_i = \mathcal{F}_{\mathrm{pre}}(x_i;\, \mathcal{C}_{<i}^{\mathrm{pre}})$. Full-depth
inference initializes a latent state $s_i^{(0)}$ from Huginn's standard truncated Gaussian
state initializer (seeded deterministically at test time) and iterates $k = 1, \dots, T$ ($T = 32$) via a
learned linear adapter $W_{\mathrm{adap}}$:
\begin{align}
  s_i^{(k)} &= \mathcal{F}_{\mathrm{core}}\!\left(
    W_{\mathrm{adap}}\!\left[s_i^{(k-1)};\, u_i\right];\,
    \mathcal{C}_{<i}^{(\mathrm{core},\, k)}
  \right), \nonumber \\
  p(x_{i+1} \mid x_{\le i}) &= \mathrm{softmax}\!\left(
    W_{\mathrm{out}}\, \mathrm{Norm}\!\left(
      \mathcal{F}_{\mathrm{coda}}\!\left(\mathrm{Norm}(s_i^{(T)});\, \mathcal{C}_{<i}^{\mathrm{coda}}\right)
    \right)
  \right).
  \label{eq:huginn}
\end{align}
Under late-entry block decoding with budget $K < T$, the active $(1 + B)^2$ block
executes Prelude once (writing exact Prelude KV states for all committed tokens, so
Prelude is \emph{never} recomputed during catch-up), initializes
$s_i^{(T-K)} = s_i^{(0)}$, runs only the final $K$ Core steps $k = T - K + 1, \dots, T$, and
reads out through Coda ($4 + 4K$ serial layer calls vs.\ $132$ at full depth). Across
the next $\lceil D_{\mathrm{bg}} / K \rceil$ steps (default full-depth $D_{\mathrm{bg}}
= T$, or an optional partial-cache approximation completing only the final
$D_{\mathrm{bg}} < T$ Core steps), committed groups advance at $K$ steps per call
alongside the active block, fusing Coda refresh into the final catch-up call at zero
extra serial forward calls. To ensure hardware-matched
wall-clock comparisons across sequential decoding, late entry, synchronous repair, and
early exit, all Huginn decoders run inside a unified CUDA-graphed paged
FlashAttention-3 engine that packs the $(1 + B)^2$ active
block and all in-flight catch-up stages into a single fixed-shape paged attention call
per layer with fused RoPE and RMSNorm kernels. For two-stage adaptation
on Dolci-Instruct-SFT, Stage~1 runs one epoch of full-depth ($T = 32$) SFT and Stage~2
runs one epoch ($7{,}406$ optimization steps, $\approx 1.9$B tokens per stage or
$< 0.25\%$ of \texttt{huginn-0125}'s $800$B-token pretraining budget, adding zero extra
inference parameters) of MARS block fine-tuning at global batch size $256$ in
\texttt{bf16}, sequence length $1{,}024$, learning rate $10^{-5}$, and truncated
backpropagation through the final $8$ recurrent steps matching \citet{geiping2025scaling}: $50\%$ of blocks execute all $32$
Core passes, while $50\%$ execute $K \sim \mathrm{Uniform}\{2, 4, 6, 8, 12, 16, 24\}$
passes (final $K$ passes for late entry; initial $K$ passes for the matched
early-exit control).

\paragraph{Single-token curves, matched-throughput frontier, and cache-completion lag ablation.}
Table~\ref{tab:hgsingle} reports full-set ($n = 1{,}319$) single-token sequential
GSM8K curves on our adapted checkpoint: running $K = 8$ final Core passes over a
completed $32$-pass cache matches full $32$-pass accuracy identically ($.396$ vs.\
$.396$, $.989$ token agreement), beating $8$-pass early exit over a $32$-pass oracle
cache ($.288$) by \textbf{+10.8 points} and $8$-pass shallow truncation ($.263$) by
\textbf{+13.3 points} ($+27.7$ points at $K = 4$). Table~\ref{tab:hgcontest} compares
our matched late-entry and early-exit Stage-2 checkpoints ($7{,}406$
steps each) at $B = 8$ across GSM8K ($K \in \{8, 12, 16, 32\}$, spanning $25\%$ to
$100\%$ Core depth in exact correspondence with $K \in \{1, 2, 3, 4\}$ on $T = 4$ Ouro)
and IFEval (across all four capability-active benchmarks, intermediate $K = 12$ late
entry reaches $6.44\times$ at $.545$ on IFEval, $7.27\times$ at $.395$ on GSM8K,
$8.20\times$ at $.281$ on BBH, and $9.75\times$ at $.234$ on MBPP). Because paged
FlashAttention-3 eliminates multi-segment dispatch overhead, pipelined catch-up adds
only $7\%$ to $12\%$ latency at equal $K$ and outperforms trained early exit across all
matched-throughput GSM8K operating points in Table~\ref{tab:hgcontest}: $K = 12$ late
entry ($.395$ at $157.8$~tokens/s) is both faster and $+2.4$ points higher than
$16$-pass early exit ($.372$ at $140.3$~tokens/s), $K = 8$ late entry ($.354$ at
$198.1$~tokens/s) is faster and higher than $12$-pass early exit ($.342$ at
$174.4$~tokens/s), and our two-stage partial-cache $K = 8$ variant ($.369$ at
$215.5$~tokens/s) beats $8$-pass early exit ($.271$ at $222.9$~tokens/s) by
\textbf{+9.7 points} ($p = 3.4\times10^{-14}$). On IFEval, the two checkpoints do not
differ significantly at $K \in \{12, 16, 32\}$ ($p \ge .30$), while above $200$~tokens/s
late entry retains full-depth accuracy down to $12.5\%$ Core depth and leads trained
early exit by \textbf{+4.8 points} at matched $\sim 225$~tokens/s ($K = 6$ at $.560$,
$223.0$~tokens/s vs.\ $K = 8$ at $.512$, $226.6$~tokens/s, $p = .029$), \textbf{+7.0
points} at $K = 6$ ($p = .0018$), and \textbf{+12.4 points} at $K = 4$ ($p =
3.0\times10^{-8}$). Even on the \emph{early-exit-trained} checkpoint itself (which never
saw late entry in Stage~2), Prelude's normalized anchor $u_i$ allows $K = 8$ late entry
to beat that checkpoint's own trained $K = 8$ early exit on GSM8K ($.327$ vs.\ $.271$ at
$193.9$~tokens/s, $p = 2.2\times10^{-6}$). Finally, Table~\ref{tab:hgcatchup}
characterizes when provisional KV states degrade at $T = 32, K = 8, B = 8$ by comparing
full-depth completion ($D_{\mathrm{bg}} = 32$, $4$ stages, $3$-block lag) against
partial-cache approximations ($D_{\mathrm{bg}} \in \{24, 16\}$, $3$ and $2$ stages,
which leave early passes $1 \to 32 - D_{\mathrm{bg}}$ uncomputed in exchange for $2$-
and $1$-block lag). On GSM8K, where each chain-of-thought arithmetic step reads digits
emitted in the immediately preceding $\sim 10$ tokens, $4$-stage full completion trails
CUDA-graphed synchronous repair by $3.3$ points ($.354$ vs.\ $.387$, $p = .0013$), and
accuracy varies monotonically with the local lag window ($.387 \to .369 \to .360 \to
.354 \to .331$ across $0, 1, 2, 3, \infty$ block lag). The $2$-stage partial-cache
approximation ($17 \to 32$, one-block lag $\approx 3.5$ tokens) narrows this gap to
$1.9$ points ($.369$ vs.\ $.387$, $p = .065$; $+3.7$ points over omitting catch-up at
$p = 1.8\times10^{-4}$) while running \textbf{3.66$\times$ faster} than synchronous
repair ($215.5$ vs.\ $58.9$~tokens/s). By contrast, on IFEval, where structural and
formatting constraints depend weakly on local provisional refinements, omitting or
lagging cache completion leaves accuracy virtually unchanged ($.553$ at no catch-up and
two-stage partial vs.\ $.551$ sequential full depth), even though early exit still
collapses at small $K$ because the active token itself requires late Core passes.

\begin{table}[H]
\caption{\textbf{Single-token sequential GSM8K evaluation on Huginn-3.5B}
($n = 1{,}319$, batched greedy decoding; gaps computed prior to rounding). ``Shallow
$r(K)$'' runs $K$ Core passes on both token and context; ``Oracle-cache early'' runs
initial passes $1, \dots, K$ over a $32$-pass context; ``Late entry'' runs final
passes $32 - K + 1, \dots, 32$ over a $32$-pass context (full depth $T = 32$ scores
$.396$ at $132$ serial layers/token, vs.\ $4 + 4K$ serial layers/token at budget $K$).}
\label{tab:hgsingle}
\vspace{-5pt}
\begin{center}
\small
\renewcommand{\arraystretch}{0.90}
\shrinktofit{%
\begin{tabular}{lcccccc}
\toprule
Core budget & Serial layers / tok & Shallow $r(K)$ & Oracle-cache early & Late entry & Token agree.\ w/\ $T\!=\!32$ & Late vs.\ $r(K)$ ($p$) \\
\midrule
$K = 4$  & 20\;\;(6.6$\times$) & .039 & .043 & \textbf{.315} & .957 & \textbf{+27.7} ($8.7\times10^{-92}$) \\
$K = 8$  & 36\;\;(3.7$\times$) & .263 & .288 & \textbf{.396} & .989 & \textbf{+13.3} ($9.8\times10^{-25}$) \\
$K = 12$ & 52\;\;(2.5$\times$) & .350 & .368 & \textbf{.393} & .995 & \textbf{+4.3} ($4.5\times10^{-5}$) \\
$K = 16$ & 68\;\;(1.9$\times$) & .378 & .383 & \textbf{.400} & .997 & \textbf{+2.2} ($.020$) \\
\bottomrule
\end{tabular}}
\end{center}
\vspace{-9pt}
\end{table}

\begin{table}[H]
\caption{\textbf{Trained early exit versus late entry on Huginn-3.5B
($T = 32$) under matched Stage-2 fine-tuning} inside the unified CUDA-graphed paged
FlashAttention-3 engine at block size $B = 8$ across full GSM8K ($n = 1{,}319$) and
strict IFEval ($n = 541$), following Table~\ref{tab:contest} ($\dagger$ marks paired
sign-test $p < .05$; gaps computed prior to rounding; ``$K = 8$ ($2$-stg partial)''
completes passes $17 \to 32$ and is compared against $8$-pass early exit at matched
$\sim 220$~tokens/s).}
\label{tab:hgcontest}
\vspace{-5pt}
\begin{center}
\small
\renewcommand{\arraystretch}{0.92}
\setlength{\tabcolsep}{5pt}
\begin{tabular}{lcccccc}
\toprule
& \multicolumn{2}{c}{\textbf{Early exit, trained}} & \multicolumn{2}{c}{\textbf{Late entry, trained}} & & \\
\cmidrule(lr){2-3}\cmidrule(lr){4-5}
Core passes / token & Acc. & tok/s & Acc. & tok/s & Gap (pts) & Sign-test $p$ \\
\midrule
\multicolumn{7}{l}{\emph{GSM8K} (1{,}319 test problems; full-depth sequential: .393 at 21.7 tok/s)} \\
$K = 8$ ($4$-stg full)     & .271 & 223 & \textbf{.354} & 198 & \textbf{+8.3}$^\dagger$  & $3.1\times10^{-10}$ \\
$K = 8$ ($2$-stg partial)  & .271 & 223 & \textbf{.369} & 215 & \textbf{+9.7}$^\dagger$  & $3.4\times10^{-14}$ \\
$K = 12$                   & .342 & 174 & \textbf{.395} & 158 & \textbf{+5.3}$^\dagger$  & $7.5\times10^{-5}$ \\
$K = 16$                   & .372 & 140 & \textbf{.405} & 127 & \textbf{+3.3}$^\dagger$  & $.015$ \\
$K = 32$ (full)            & .382 &  73 & \textbf{.397} &  73 & \textbf{+1.5}\phantom{$^\dagger$} & $.28$ \\
\midrule
\multicolumn{7}{l}{\emph{IFEval} (541 test prompts; full-depth sequential: .551 at 21.6 tok/s)} \\
$K = 4$                    & .407 & 364 & \textbf{.530} & 268 & \textbf{+12.4}$^\dagger$ & $3.0\times10^{-8}$ \\
$K = 6$                    & .490 & 284 & \textbf{.560} & 223 & \textbf{+7.0}$^\dagger$  & $.0018$ \\
$K = 8$ ($4$-stg full)     & .512 & 227 & \textbf{.542} & 193 & \textbf{+3.0}\phantom{$^\dagger$} & $.16$ \\
$K = 8$ ($2$-stg partial)  & .512 & 227 & \textbf{.553} & 205 & \textbf{+4.1}\phantom{$^\dagger$} & $.053$ \\
$K = 12$                   & \textbf{.567} & 158 & .545 & 139 & $-$2.2\phantom{$^\dagger$} & $.30$ \\
$K = 16$                   & .549 & 119 & \textbf{.566} & 113 & \textbf{+1.7}\phantom{$^\dagger$} & $.48$ \\
$K = 32$ (full)            & .543 &  62 & \textbf{.551} &  63 & \textbf{+0.7}\phantom{$^\dagger$} & $.79$ \\
\bottomrule
\end{tabular}
\end{center}
\vspace{-6pt}
\end{table}

\begin{table}[H]
\caption{\textbf{Cache-completion lag and depth ablation on Huginn-3.5B} at $T = 32,
K = 8, B = 8$ across full GSM8K ($n = 1{,}319$; sequential $T = 32$ is $.393$ at
$21.7$~tokens/s) and strict IFEval ($n = 541$; sequential $T = 32$ is $.551$ at
$21.6$~tokens/s; gaps prior to rounding). Unlike $4$-stage full completion ($1 \to 32$),
the $2$- and $3$-stage policies are partial-cache approximations that omit early passes
to shrink the provisional lag window, narrowing the GSM8K gap against synchronous repair
from $3.3$ to $1.9$ points while increasing throughput.}
\label{tab:hgcatchup}
\vspace{-3pt}
\begin{center}
\small
\renewcommand{\arraystretch}{0.94}
\shrinktofit{%
\begin{tabular}{lcccccc}
\toprule
& & & \multicolumn{2}{c}{\textbf{GSM8K} ($n\!=\!1{,}319$)} & \multicolumn{2}{c}{\textbf{IFEval} ($n\!=\!541$)} \\
\cmidrule(lr){4-5}\cmidrule(lr){6-7}
Cache-completion policy ($K = 8$ of $32$) & Stages (lag) & $1\!+\!\mathbb{E}[c]$ & tok/s (speedup) & Acc. & tok/s (speedup) & Acc. \\
\midrule
Synchronous (blocking $1 \to 32$, CUDA-graphed) & $0$ ($0$ blks) & 3.61 & \phantom{0}58.9 \ (2.71$\times$) & .387 & n/a & n/a \\
2-stage partial (passes $17 \to 32$)            & $2$ ($1$ blk)  & 3.52 & \textbf{215.5 \ (9.93$\times$)} & \textbf{.369} & \textbf{205.0 \ (9.50$\times$)} & \textbf{.553} \\
3-stage partial (passes $9 \to 32$)             & $3$ ($2$ blks) & 3.52 & 205.1 \ (9.45$\times$) & .360 & 191.5 \ (8.88$\times$) & .538 \\
4-stage full (passes $1 \to 32$, default)       & $4$ ($3$ blks) & 3.51 & 198.1 \ (9.13$\times$) & .354 & 193.3 \ (8.96$\times$) & .542 \\
Never (keep provisional $8$-pass cache)         & n/a ($\infty$) & 3.45 & 231.6 \ (10.68$\times$) & .331 & 228.0 \ (10.57$\times$) & .553 \\
\bottomrule
\end{tabular}}
\end{center}
\vspace{-4pt}
\end{table}

\section{Scaling Across Request Concurrency}
\label{app:concurrency}

\begin{figure}[!htbp]
\begin{center}
\includegraphics[width=0.98\textwidth]{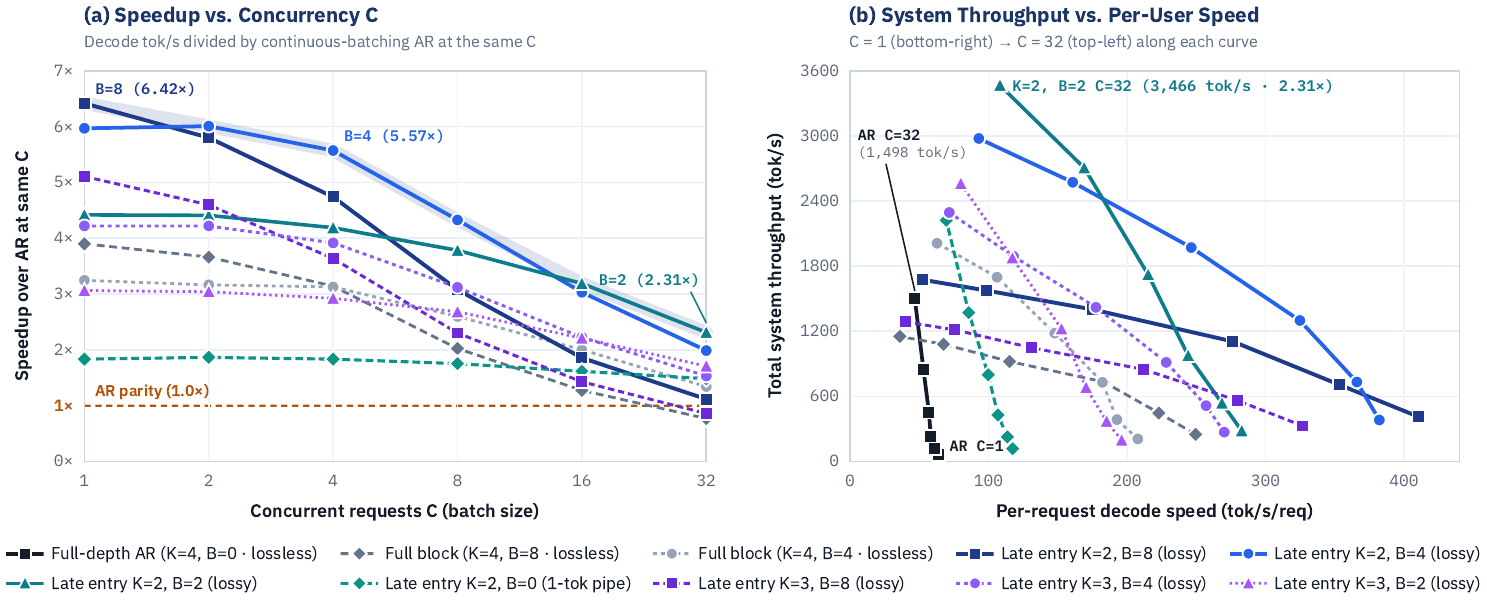}
\end{center}
\vspace{-6pt}
\caption{\textbf{Steady-state concurrency scaling on Ouro-2.6B ($T = 4$, GSM8K, single
H100 GPU)} across $C \in \{1, 2, 4, 8, 16, 32\}$ concurrent requests.
\textbf{(a)~Speedup over continuous-batching AR at the same $C$:} fixed large blocks
($B = 8$) decay steeply as $C \times 99$ rows saturate compute, whereas selecting
smaller discrete block profiles with load ($B \in \{8, 4, 2\}$, shaded $K = 2$ profile
envelope) sustains \textbf{2.31$\times$} at $C = 32$ ($K = 2$) and \textbf{1.71$\times$}
($K = 3$).
\textbf{(b)~System decode throughput vs.\ per-request speed} ($C = 1$ at bottom-right
to $C = 32$ at top-left along each curve).}
\label{fig:concurrency}
\vspace{-4pt}
\end{figure}

How does late-entry block decoding behave when serving $C > 1$ concurrent requests on
a single GPU? To characterize the transition from memory-bandwidth-bound to
compute-bound execution, we extend our paged FlashAttention-3
\citep{shah2024flashattention3} / CUDA-graph engine on
Ouro-2.6B ($T = 4$) to multi-request continuous batching across active concurrencies
$C \in \{1, 2, 4, 8, 16, 32\}$ on GSM8K (single NVIDIA H100 80GB GPU, \texttt{bf16},
\texttt{max\_new}$\,= 1{,}024$, dynamic paged KV slot allocation with immediate slot
refill upon request completion, and zero token discrepancies between CUDA-graph replay
and eager execution across all eleven modes). At every concurrency $C$, we normalize
against full-depth ($T = 4$) continuous-batching autoregressive decoding (\texttt{AR},
$K = 4, B = 0$) at the \emph{same} $C$, reporting both pure decode throughput and
end-to-end serving throughput including online four-pass prefills for newly arriving
prompts (Figure~\ref{fig:concurrency} and Table~\ref{tab:concurrency}).

Figure~\ref{fig:concurrency} and Table~\ref{tab:concurrency} highlight three scaling
behaviors. First, \textbf{speedup compression at fixed $B = 8$ comes from $(1 + B)^2$
block expansion rather than late entry}: as concurrency grows from $C = 1$ to $C = 32$,
lossless $K = 4, B = 8$ block decoding collapses $5.1\times$ ($3.90\times \to
0.77\times$), whereas late entry's multiplicative gain over same-$B$ full-depth block
decoding ($K = 2$ over $K = 4$, middle block of Table~\ref{tab:concurrency}) stays
nearly invariant at \textbf{1.65$\times \to$ 1.45$\times$} at $B = 8$ and
\textbf{1.84$\times \to$ 1.48$\times$} at $B = 4$ ($1.30\times \to 1.14\times$ for
$K = 3$). Physically, $K = 4, B = 8$ runs all four passes across all $(1 + 8)^2 = 81$
candidate rows ($324$ row-passes/step), whereas $K = 2, B = 8$ runs only two passes on
the $81$ block rows and restricts catch-up passes to the $\le 9$ committed tokens
($198$ row-passes/step, \textbf{39\% fewer FLOPs per step}), preserving late entry's
advantage over full-depth block decoding even when Tensor Cores saturate. Second,
\textbf{selecting smaller discrete block profiles $B \in \{8, 4, 2\}$ as concurrency
rises shifts the crossover substantially} (shaded profile envelope in
Figure~\ref{fig:concurrency}a; as in standard multi-bucket serving engines,
pre-capturing one CUDA graph per discrete $(B, C)$ bucket shares the same model weights
and paged KV pool, allowing $O(1)$ per-step profile selection from active batch size $C$
without graph recompilation): $B = 8$ is fastest at $C = 1$, $B = 4$ ($35$ rows/request) takes over across $C = 2$
to $8$ ($6.01\times$ to $4.33\times$ at $K = 2$; $4.22\times$ to $3.12\times$ at
$K = 3$), and $B = 2$ ($15$ rows/request) leads at $C = 16$ to $32$, reaching
\textbf{3,466~tokens/s} (\textbf{2.31$\times$} decode; $108.3$~tokens/s per request,
exceeding $C = 1$ sequential \texttt{AR} at $64.0$~tokens/s) at $K = 2$ and
\textbf{2,561~tokens/s} (\textbf{1.71$\times$}) at $K = 3$ (vs.\ $1.34\times$ for
lossless $K = 4, B = 4$), with \textbf{1.67$\times$} and \textbf{1.47$\times$}
end-to-end throughput at $C = 32$ including online four-pass prefills. Third,
\textbf{single-token pipelined late entry ($B = 0$)} uses zero draft branches ($3$
rows/request/call at $K = 2$, executing $T + K = 6$ total passes per token versus $4$
for plain \texttt{AR}) and sustains a flat $1.83\times \to 1.48\times$ decode speedup
across $C = 1$ to $32$.

\begin{table}[!htbp]
\vspace{-4pt}
\caption{\textbf{Multi-request continuous-batching throughput and mechanism
factorization on Ouro-2.6B ($T = 4$, GSM8K, single H100 GPU)} across concurrencies
$C \in \{1, 2, 4, 8, 16, 32\}$. Top block: total decode tokens/s (and speedup vs.\
same-$C$ \texttt{AR}); middle block: isolated multiplicative gain of late entry over
matched-$B$ full-depth ($K = 4$) block decoding; bottom block: end-to-end serving
throughput including online four-pass slot-refill prefills.}
\label{tab:concurrency}
\vspace{-5pt}
\begin{center}
\small
\renewcommand{\arraystretch}{0.90}
\shrinktofit{%
\begin{tabular}{llcccccc}
\toprule
Mode & Regime (rows/req/call) & $C = 1$ & $C = 2$ & $C = 4$ & $C = 8$ & $C = 16$ & $C = 32$ \\
\midrule
\multicolumn{8}{l}{\emph{Pure decode throughput (total decode tokens/s and speedup vs.\ same-$C$ \textup{\texttt{AR}})}} \\
\texttt{AR} ($K\!=\!4, B\!=\!0$) & Lossless ref.\ ($1$) & 64.0 (1.00$\times$) & 121.8 (1.00$\times$) & 233.2 (1.00$\times$) & 455.4 (1.00$\times$) & 848.6 (1.00$\times$) & 1{,}498.2 (1.00$\times$) \\
$K = 4, B = 8$ & Lossless ref.\ ($81$) & 249.6 (3.90$\times$) & 446.0 (3.66$\times$) & 732.9 (3.14$\times$) & 921.5 (2.02$\times$) & 1{,}080.1 (1.27$\times$) & 1{,}152.4 (0.77$\times$) \\
$K = 4, B = 4$ & Lossless ref.\ ($25$) & 207.8 (3.25$\times$) & 385.5 (3.17$\times$) & 729.1 (3.13$\times$) & 1{,}182.7 (2.60$\times$) & 1{,}698.3 (2.00$\times$) & 2{,}011.2 (1.34$\times$) \\
$K = 3, B = 8$ & Lossy late ($99/90/90$) & \textbf{327.0 (5.11$\times$)} & \textbf{560.1 (4.60$\times$)} & 848.0 (3.64$\times$) & 1{,}048.3 (2.30$\times$) & 1{,}212.6 (1.43$\times$) & 1{,}285.8 (0.86$\times$) \\
$K = 3, B = 4$ & Lossy late ($35/30/30$) & 270.2 (4.22$\times$) & 514.1 (4.22$\times$) & \textbf{913.3 (3.92$\times$)} & \textbf{1{,}420.6 (3.12$\times$)} & \textbf{1{,}890.0 (2.23$\times$)} & 2{,}294.5 (1.53$\times$) \\
$K = 3, B = 2$ & Lossy late ($15/12/12$) & 196.2 (3.06$\times$) & 370.6 (3.04$\times$) & 681.8 (2.92$\times$) & 1{,}221.7 (2.68$\times$) & 1{,}875.7 (2.21$\times$) & \textbf{2{,}560.8 (1.71$\times$)} \\
$K = 3, B = 0$ & Lossy 1-tok ($3/2/2$)   & 79.6 (1.24$\times$)  & 153.0 (1.26$\times$) & 293.6 (1.26$\times$) & 545.7 (1.20$\times$)   & 952.3 (1.12$\times$)   & 1{,}548.7 (1.03$\times$) \\
$K = 2, B = 8$ & Lossy late ($99$)       & \textbf{410.8 (6.42$\times$)} & 706.8 (5.80$\times$) & 1{,}105.7 (4.74$\times$) & 1{,}400.9 (3.08$\times$) & 1{,}575.7 (1.86$\times$) & 1{,}674.5 (1.12$\times$) \\
$K = 2, B = 4$ & Lossy late ($35$)       & 382.2 (5.97$\times$) & \textbf{732.0 (6.01$\times$)} & \textbf{1{,}299.4 (5.57$\times$)} & \textbf{1{,}971.0 (4.33$\times$)} & 2{,}573.0 (3.03$\times$) & 2{,}976.8 (1.99$\times$) \\
$K = 2, B = 2$ & Lossy late ($15$)       & 282.8 (4.42$\times$) & 537.1 (4.41$\times$) & 976.6 (4.19$\times$)   & 1{,}722.3 (3.78$\times$) & \textbf{2{,}705.5 (3.19$\times$)} & \textbf{3{,}466.0 (2.31$\times$)} \\
$K = 2, B = 0$ & Lossy 1-tok ($3$)       & 117.5 (1.83$\times$) & 227.3 (1.87$\times$) & 427.4 (1.83$\times$)   & 798.5 (1.75$\times$)   & 1{,}371.6 (1.62$\times$) & 2{,}222.9 (1.48$\times$) \\
\midrule
\multicolumn{8}{l}{\emph{Isolated late-entry multiplier over same-$B$ full-depth ($K = 4$) block decoding}} \\
$K = 2 \;/\; K = 4$ at $B = 8$ & $198$ vs.\ $324$ row-passes/step & \textbf{1.65$\times$} (+65\%) & \textbf{1.58$\times$} (+58\%) & \textbf{1.51$\times$} (+51\%) & \textbf{1.52$\times$} (+52\%) & \textbf{1.46$\times$} (+46\%) & \textbf{1.45$\times$} (+45\%) \\
$K = 2 \;/\; K = 4$ at $B = 4$ & $70$ vs.\ $100$ row-passes/step  & \textbf{1.84$\times$} (+84\%) & \textbf{1.90$\times$} (+90\%) & \textbf{1.78$\times$} (+78\%) & \textbf{1.67$\times$} (+67\%) & \textbf{1.52$\times$} (+52\%) & \textbf{1.48$\times$} (+48\%) \\
$K = 3 \;/\; K = 4$ at $B = 4$ & $95$ vs.\ $100$ row-passes/step  & 1.30$\times$ (+30\%) & 1.33$\times$ (+33\%) & 1.25$\times$ (+25\%) & 1.20$\times$ (+20\%) & 1.11$\times$ (+11\%) & 1.14$\times$ (+14\%) \\
\midrule
\multicolumn{8}{l}{\emph{End-to-end serving throughput including online 4-pass slot-refill prefills (total e2e tokens/s)}} \\
\texttt{AR} ($K\!=\!4, B\!=\!0$) & Lossless ref. & 61.5 (1.00$\times$) & 113.2 (1.00$\times$) & 200.3 (1.00$\times$) & 348.6 (1.00$\times$) & 539.6 (1.00$\times$) & 803.5 (1.00$\times$) \\
$K = 4, B = 4$ & Lossless ref. & 183.2 (2.98$\times$) & 312.0 (2.76$\times$) & 501.4 (2.50$\times$) & 648.5 (1.86$\times$) & 880.7 (1.63$\times$) & 1{,}100.0 (1.37$\times$) \\
$K = 3, B = 4$ & Lossy late    & 229.0 (3.72$\times$) & 388.9 (3.44$\times$) & 573.2 (2.86$\times$) & 711.2 (2.04$\times$) & 946.0 (1.75$\times$) & \textbf{1{,}185.1 (1.47$\times$)} \\
$K = 2, B = 4$ & Lossy late    & 305.8 (4.97$\times$) & 501.8 (4.43$\times$) & \textbf{699.0 (3.49$\times$)} & \textbf{844.1 (2.42$\times$)} & \textbf{1{,}080.6 (2.00$\times$)} & \textbf{1{,}339.0 (1.67$\times$)} \\
\bottomrule
\end{tabular}}
\end{center}
\vspace{-4pt}
\end{table}

\end{document}